\documentclass[journal]{IEEEtran}
\usepackage{amsmath,amsfonts}
\usepackage{algorithmic}
\usepackage{algorithm}
\usepackage{array}
\usepackage[caption=false,font=normalsize,labelfont=sf,textfont=sf]{subfig}
\usepackage{textcomp}
\usepackage{stfloats}
\usepackage{url}
\usepackage{verbatim}
\usepackage{graphicx}
\usepackage{cite}
\usepackage{booktabs}
\usepackage{hyperref}

\begin{document}

\title{Second-order Gaussian directional derivative representations for image high-resolution corner detection}
	
\author{Jiamiao Lu$^{\dagger}$, Junjie Qiu$^{\dagger}$, Dongbo Xie$^{\dagger}$, Lingkun Ma, Changming Sun, Weichuan Zhang

\IEEEcompsocitemizethanks{
	\IEEEcompsocthanksitem $^{\dagger}$These authors contributed equally.
	\IEEEcompsocthanksitem J. Lu, J. Qiu,  D. Xie, L. Ma and W. Zhang are with School of Electronic Information and Artificial Intelligence, Shaanxi University of Science and Technology, Xi'an, Shaanxi Province, China.\protect\\
E-mails:~241612058@sust.edu.cn, 241612105@sust.edu.cn; Junjie.Qiu@sust.edu.cn, malingkun@sust.edu.cn, zwc2003@163.com;
	\IEEEcompsocthanksitem C. Sun is with CSIRO Data61, PO Box 76, Epping, NSW 1710, Australia.\protect\\
	E-mail: changming.sun@csiro.au\\
	Corresponding author: Weichuan Zhang.
}
}

\maketitle

\begin{abstract}
Corner detection is widely used in various computer vision tasks, such as image matching and 3D reconstruction. Our research indicates that there are theoretical flaws in Zhang et al.'s use of a simple corner model to obtain a series of corner characteristics, as the grayscale information of two adjacent corners can affect each other. In order to address the above issues, a second-order Gaussian directional derivative (SOGDD) filter is used in this work to smooth two typical high-resolution angle models (i.e. END-type and L-type models). Then, the SOGDD representations of these two corner models were derived separately, and many characteristics of high-resolution corners were discovered, which enabled us to demonstrate how to select Gaussian filtering scales to obtain intensity variation information from images, accurately depicting adjacent corners. In addition, a new high-resolution corner detection method for images has been proposed for the first time, which can accurately detect adjacent corner points. The experimental results have verified that the proposed method outperforms state-of-the-art methods in terms of localization error, robustness to image blur transformation, image matching, and 3D reconstruction.
\end{abstract}


\section{Introduction}
\IEEEPARstart{C}{orner} detection is highly applicable in various computer vision tasks such as image matching and 3D reconstruction~\cite{jing2022image}. Existing image corner detection methods can be roughly divided into two categories: handcrafted and deep learning-based methods. Handcrafted methods~\cite{mair2010adaptive,lowe2004distinctive,lucas1981iterative} rely on predefined rules or mathematical models for identifying corners from images. Deep learning-based methods~\cite{trujillo2006synthesis,zhao2023deep,law2018cornernet,jing2023ecfrnet, wang2020corner,wang2018survey,li2019multi,liu2024aekan,ren2024few,zhang2021ndpnet,li2023mutual,gao2020fast,jing2021novel,lu2022image,bao2022corner} utilize convolutional neural networks (CNNs)~\cite{law2018cornernet,liu2016ssd} or Transformer~\cite{vaswani2017} as a backbone~\cite{carion2020end} to learn feature representations for extracting corners from images.

\begin{figure}[!t]
	\centering
	\includegraphics[width=3.5in]{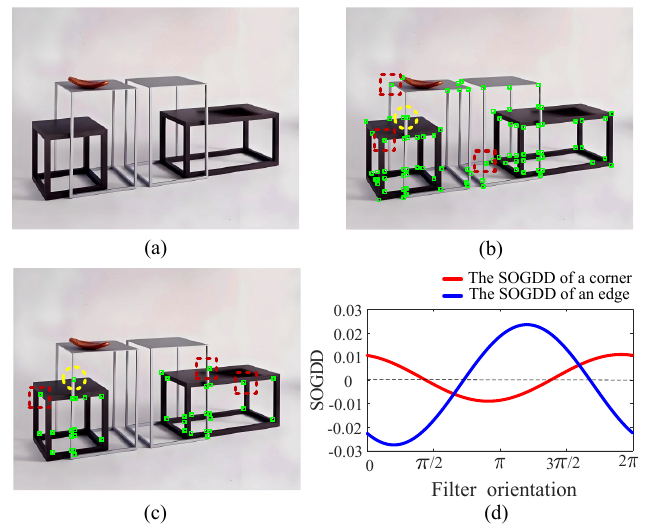}
	\caption{The results of image corner detection by the SOGDD detector~\cite{zhang2021corner} with different smoothing scales. (a) An image; (b) The results of corner detection by the SOGDD detector~\cite{zhang2021corner} with a given Gaussian smoothing scale ($\sigma^{2}$=2); (c) The results of corner detection by the SOGDD detector~\cite{zhang2021corner} with another Gaussian smoothing scale ($\sigma^{2}$=5); (d) The SOGDD of a corner and an edge point.}
	\label{fig1}
\end{figure}

It was indicated in~\cite{zhang2021corner,zhang2020corner,zhang2023image,zhang2019discrete}, corners should always be detected under different filtering scales. Take an image as shown in Fig.~\ref{fig1}(a) as an example, the corner detection results of the image under different Gaussian scale filtering by the second-order Gaussian directional derivative (SOGDD) filters~\cite{zhang2021corner} are shown in Figs.~\ref{fig1}(b) and (c). It can be observed from Fig.~\ref{fig1}(b) that some adjacent corners (marked by `$\square$') cannot be accurately detected by the SOGDD detector~\cite{zhang2021corner} with a given Gaussian scale ($\sigma^{2}$=2). Meanwhile, a pair of adjacent corners (marked by `$\bigcirc$') are detected by the SOGDD detector~\cite{zhang2021corner} with the given Gaussian scale as shown in Fig.~\ref{fig1}(b). In contrast, this pair of adjacent corners cannot be accurately detected and an edge (marked by `$\bigcirc$') between them is detected as a corner by the SOGDD detector~\cite{zhang2021corner} with another Gaussian scale ($\sigma^{2}$=5) as shown in Fig.~\ref{fig1}(c). This is because the distance between two adjacent corners is too close. When the SOGDD filters with different scales smooth them, it may cause their corresponding SOGDD to interfere with each other. Take the left corner (marked by `$\bigcirc$') as shown in Fig.~\ref{fig1}(b) and an edge (marked by `$\bigcirc$') as shown in Fig.~\ref{fig1}(c) as an example, their corresponding SOGDDs with a Gaussian scale ($\sigma^{2}$=5) are shown in Fig.~\ref{fig1}(d). It can be found that the SOGDDs of the edge are larger than the SOGDDs of the corner in multiple directions, which results in the inability to accurately detect adjacent corners as shown in Fig.~\ref{fig1}(b), and even incorrect corner detection as shown in Fig.~\ref{fig1}(c). The above experimental phenomena triggered our following thoughts: When two corners are close to each other, how to filter them and accurately obtain their corresponding intensity variation information for detecting the adjacent corners (i.e., image high resolution corner detection)?

To address the aforementioned problems, in this work, the contour-based END-type corner model and L-type corner model~\cite{zhang2019discrete, jing2022recent, pan2024pseudo, jing2022image, zhang2024re,6507646,shui2012noise, zhang2017noise, zhang2015contour, zhang2020corner, zhang2019discrete, zhang2014corner, zhang2019corner, li2023traffic, qiu2021recurrent, zhang2023image,wang2024unbiased,lei2024semi,liao2022asrsnet,islam2023background, wang2025principal,zheng2023fully,lu2023track,liao2025dynamic,ren4962361adaptive,an2023edge} are extended to the gray-value-based END-type corner model and L-type corner model respectively, then the SOGDD filters are used to smooth the two high-resolution corner models separately. The SOGDD representations of the two types of corner models are derived, which will help us to demonstrate how to select Gaussian scale to accurately obtain intensity variation information from images for depicting adjacent corners. The highlights of the proposed method are: (1)~The SOGDD representations of the gray-value-based END-type and L-type corner models are derived, which helps us how to accurately obtain intensity variation information from images for depicting adjacent corners; (2)~Some properties of high resolution corners are summarized; (3)~A new image high-resolution corner detection method is proposed which has the capability to accurately detect high-resolution corners. Furthermore, experimental results verify the superiority of the proposed method in terms of localization error, robustness to different image afﬁne transformations, image matching, and 3D reconstruction. 

The rest of this paper is organized as follows. In Section~2, the background and related works are briefly introduced. The SOGDD representations of image high resolution corners are derived, some properties of high resolution corners are summarized, and a new image high resolution corner detection method are presented in Section 3. Experimental results are shown and analyzed in Section 4. Finally, conclusions are drawn in Section 5.
 
\section{Related Work}
The existing image variation information-based interest points (including corners and blobs) detection methods, which can be classified into two groups: handcrafted and deep learning-based image interest point detection methods, are briefly reviewed.

Harris and Stephens~\cite{harris1988combined} developed the Plessey detector which utilizes the first-order image variation information along horizontal and vertical directions to construct a 2$\times$2 structure tensor for detecting corners from images. Smith~\cite{smith1997susan} proposed the smallest univalve segment assisting nucleus (SUSAN) detector, which uses the difference in grayscale values between pixels in the area covered by the template and those at the center point for finding corners. Lowe~\cite{lowe2004distinctive} proposed the scale-invariant feature transform (SIFT) interest points detection method, Gaussian difference was used to extract candidate interest points from images in scale space, and calculate their principal curvatures using the Hessian matrix for determining whether they are interest points or not. Rosten et al.~\cite{rosten2008faster} proposed a method for image corner detection method called scale-invariant feature transform (FAST). If a pixel is surrounded by a certain number of pixels with different pixel values, it is considered a corner. Bay et al.~\cite{bay2006surf} used a box filter instead of a Gaussian filter to approximate the determinant of the Hessian matrix and extract interest points. Zhang et al.~\cite{zhang2020corner} develop corner detection metrics utilizing the eigenvalues of multi-scale and multi-directional structure tensors. And~\cite{zhang2021corner}  smooth the input image using first-order and second-order generalized Gaussian directional derivative (SOGGDD) filters to obtain the SOGGDD of each pixel, and the second-order directional derivative correlation matrix and its eigenvalues are used to obtain a corner metric that can describe the difference between edges and corners. Based on this, a corner detection method is proposed. In addition, the corner metric proposed in~\cite{zhang2023image} can accurately detect corners with different scale factors. Wang et al.~\cite{wang2024anisotropic} proposed a novel anisotropic scale-invariant (ASI) ellipse detection method that performs ellipse detection in a transformed image space called ellipse normalization (EN) space. The required ellipses in the original image are "normalized" to unit circles, proving that each ellipse can be detected with the same accuracy regardless of its size and ellipticity. Bellavia et al.~\cite{bellavia2011improving} upgraded the Harris corner detector by searching for enhanced corners near the edge and detecting corners through z-score normalization. And they further improved the selection criteria for Harris Z description and proposed HarrisZ\(^+\)~\cite{bellavia2022harrisz+} for the next generation image matching pipeline. Compared with HarrisZ detector~\cite{bellavia2011improving}, HarrisZ\(^+\) provides more but discriminative key points, which are better distributed on the image and have higher positioning accuracy.

Ono et al.~\cite{ono2018lf} proposed LF-Net for extracting interest points from images, in which the designed network has the ability to generate a scale space fractional map and dense direction estimation for selecting keypoints. Detone et al.~\cite{detone2018superpoint} proposed the SuperPoint keypoint detection method, which utilizes CNN to achieve joint training of feature point detection and descriptors and generates interest point detection in a single forward pass through a self-supervised learning method. Dusmanu et al.~\cite{dusmanu2019d2} proposed a D2-Net network for obtaining robust local feature representations from images through end-to-end training and extracting interest points under different imaging conditions, such as day-night
illumination changes~\cite{sattler2018benchmarking} and weakly textured scenes ~\cite{taira2018inloc}. Related et al.~\cite{revaud2019r2d2} proposed the R2D2 network, which leverages gradient information and confidence maps derived from feature maps to select stable and highly discriminative keypoints. Kanakis et al. ~\cite{kanakis2023zippypoint} presented fast interest point detection using a ZippyPoint network with mixed precision discretization. Zhao et al.  ~\cite{edstedt2024dedode} proposed the DeDoDe network model, a modular geometric keypoint detection method independent of descriptors, which directly learns corners from 3D consistency. Zhao et al.~\cite{zhao2024balf} proposed the BALF network for detection, which is based on an MLP network architecture. They first constructed a pure MLP-based image encoder. Subsequently, the detection module employs a softmax operator with microchannel capabilities for keypoint detection. Gao et al.~\cite{gao2023dynamic} proposed a dynamic keypoint detection network (DKDNet) that generates adaptive keypoint features through attention mechanisms and captures keypoints with different patterns. And by guiding the heatmap activator and considering the importance of different feature channels, multiple sets of keypoints are fused to achieve keypoint detection. Potje et al. ~\cite{potje2024xfeat} proposed a lightweight network architecture (XFeat) for extracting keypoints, which separates keypoint detection into an independent branch and uses 1x1 convolution to extract keypoint heatmaps on 8x8 tensor blocks. Zhao et al.~\cite{zhao2022alike} proposed that ALIKE has a Differentiable Keypoint Detection (DKD) module, which outputs accurate sub-pixel keypoints by backpropagating gradients. ALIKED~\cite{zhao2023aliked} is a further optimization of the ALIKE framework by introducing the Sparse Deformable Descriptor Head (SDDH) module, which allows the network to adaptively adjust the sampling positions around key points to detect them.

Within the scope of our investigation, the aforementioned methods have not considered how to obtain intensity variation information from images for detecting adjacent corners. In this work, we first obtain the SOGDD representations of the three types of corner models and introduce a new high-resolution image corner detection method which has the capability to obtain high-resolution image intensity variation information for accurately extracting high-resolution corners.

\section{The Proposed Method}
In this section, the SOGDD representations of different types of high-resolution corners are derived. Then the properties of high-resolution corners are summarized. Finally, a new high-resolution corner detection method is presented.

\subsection{The SOGDD Representations of Different Types of High-resolution Corners}
In Cartesian coordinates, a Gaussian filter can be represented as~\cite{zhang2023image}:
\begin{align}
	g_{\sigma}(x,y) = \frac{1}{2\pi\sigma^2} \exp\left(-\frac{(x^2+y^2)}{2\sigma^2}\right),
	\label{eq1}
\end{align}
where\( (x,y) \) is point position and \(\sigma \) is scale factor. Then the SOGDD filters can be derived as follows:
\begin{align} 
	\psi_{\sigma,\theta}(x,y) &= \frac{\partial^2 g_{\sigma}}{\partial x^2}(\mathbf{R}_{\theta}[x,y]^\tau) \notag\\
	& = \frac{1}{\sigma^2} \left( \frac{1}{\sigma^2}(x\cos\theta+y\sin\theta)^2-1 \right) g_{\sigma,\theta}(x,y),
	\label{eq2}
\end{align}
with 
\[
\mathbf{R}_{\theta}=
\begin{gathered}
	\begin{bmatrix}
		\cos\theta  &  \sin\theta \\
		-\sin\theta  &  \cos\theta 
	\end{bmatrix} ,
\end{gathered}
\]
where \( R_{\theta} \) is a rotation matrix with angle \( \theta \) and \(\tau\) represents matrix transpose.

\begin{figure}[!t]
	\includegraphics[width=3.5in]{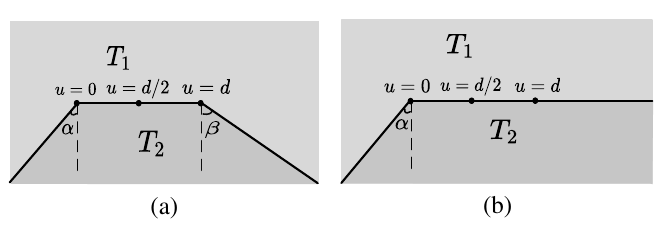}
	\caption{Two common high resolution corner models: (a) END-type high-resolution corner model; (b) L-type high-resolution corner model; \(T_1\) and \(T_2\) are grayscale values respectively, and two corners are separated by \( d \) with \(\alpha \in (0, \pi/2)\) and \(\beta \in (0, \pi/2)\), \( d \) represents the distance between two corners, an edge point locates at \( u=d/2\).}
	\label{fig_2}
\end{figure}

Than we can obtain SOGDD representation 
\begin{align}
\Psi(\theta) &=\int_{-\infty}^{+\infty} \int_{-\infty}^{+\infty} f(x,y)\psi(0-x,0-y) \,dx\,dy
\label{eq3}
\end{align}

To obtain the properties of image high-resolution corners, END-type high-resolution corner model L-type high-resolution corner model~\cite{zhang2019discrete} are constructed as shown in fig \ref{fig_2}. It can be observed that there exist two adjacent corners in the positions \(u=0\) and \(u=d\) respectively. From fig \ref{fig_2}, it can be easily seen that if the intensity variation of the two corners at the positions \(u=0\) and \(u=d\) is larger than that at the egde point \(u=d/2\), the high-resolution corners can be separated and detected.

Taking END-type corner model as follows. Corner, which locates at \( u=0 \), can be represented as:
 \begin{align}
    &f\left(x,y\right)=\begin{cases}
	T_1, & y\geq\tan\left(\frac{\pi}{2}-\alpha\right)x,x<0;\\ 
	& y\geq 0,0\leq x<d;\\ 
	& y\geq\tan\left(\frac{\pi}{2}+\beta\right)(x-d),d\leq x,\\
	T_2, & \text{otherwise}.\end{cases}
	\label{eq3}
\end{align}
The SOGDD representation \(\Psi_{1}(\theta)\) of the corner, which locates at \( u=0 \), please see equation (\ref{eq4}).
\begin{figure*}[b]
	\hrulefill
	\begin{equation}\begin{aligned}
			\Psi_{1}(\theta) &=\scalebox{1.1}{\(\int_{-\infty}^{+\infty} \int_{-\infty}^{+\infty} f(x,y)\psi(0-x,0-y) \,dx\,dy\)} \\
			&= \scalebox{1.1}{\(\left(T_1-T_2\right) \biggl(\frac{\sin^2\alpha}{2\pi\sigma^2} \biggl(\frac{\sqrt{\pi}\cos^2\theta}{2\tan\alpha} - \frac{\sin^2\theta}{\tan\alpha} - \sin2\theta\biggr)+\frac{\sin^2\beta}{2\pi\sigma^2} \exp\left(-\frac{d^2}{2\sigma^2}\right)\biggl(\sin2\theta- \frac{\sqrt{\pi}\cos^2\theta}{2\tan\beta}- \frac{\sin^2\theta}{\tan\beta}\biggr) \)} \\
			&-\scalebox{1.1}{\(\frac{\sqrt{2\pi}d\sin^3\beta}{4\pi\sigma^3} \exp\left(-\frac{d^2\cos^2\beta}{2\sigma^2}\right) \biggl[1-\Phi\left(\frac{\sqrt{2}d\sin\beta}{2\sigma}\right)\biggr] \biggl(\sin2\theta-\frac{\sqrt{\pi}\cos^2\theta}{2\tan\beta}- \frac{\sin^2\theta}{\tan\beta}\biggr)
				\)}\\
			&+\scalebox{1.1}{\(\frac{\sin2\theta}{2\pi\sigma^2} \biggl(1-\exp\left(-\frac{d^2}{2\sigma^2}\right)\biggr)+\frac{\sqrt{2\pi}d\sin\beta}{4\pi\sigma^3} \exp\left(-\frac{d^2\cos^2\beta}{2\sigma^2}\right) \biggl[1-\Phi\left(\frac{\sqrt{2}d\sin\beta}{2\sigma}\right)\biggr]\biggl(\sin2\theta+ \frac{\sqrt{\pi}\cos^2\theta}{2\tan\beta}\biggr) \biggr)\)}.
			\label{eq4}
	\end{aligned}\end{equation}
\end{figure*}

Egde point, which locates at \( u=d/2 \), can be represented: 
\begin{align}
	&f\left(x,y\right)
	=\begin{cases}T_1,
		&y\geq \tan\left(\frac{\pi}{2}-\alpha\right)(x+\frac{d}{2}), x<-\frac{d}{2};\\ & y\geq 0,-\frac{d}{2}\leq x<\frac{d}{2};\\ & y \geq\tan\left(\frac{\pi}{2}+\beta\right)(x-\frac{d}{2}), \frac{d}{2} \leq x, \\
		T_2,&\text{otherwise}.
	\end{cases}
	\label{eq5}
\end{align}

The SOGDD representation \(\Psi_{2}(\theta)\) of the edge point, which locates at \( u=d/2 \), please see equation (\ref{eq6}).
\begin{figure*}[hb]
	\begin{equation}
		\begin{aligned}
		\Psi_{2}(\theta) &=\scalebox{1.1}{\(\int_{-\infty}^{+\infty} \int_{-\infty}^{+\infty} f(x,y)\psi(0-x,0-y) \,dx\,dy\)} \\
		&= \scalebox{1.1}{\(\frac{(T_{1}-T_{2})}{2\pi\sigma^{3}} \biggl( -\frac{\sqrt{2\pi} \sin \alpha}{2} \exp\left(-\frac{d^{2} \cos^{2} \alpha}{8\sigma^{2}}\right) \biggl[ 1 - \Phi\left(\frac{\sqrt{2} d \sin \alpha}{4\sigma}\right) \biggr] \biggl( \frac{d}{2} \sin 2\theta - \frac{\sqrt{\pi} d \cos^{2} \theta}{4 \tan \alpha} \biggr)\)} \\
	   &-\scalebox{1.1}{\(\frac{\sqrt{2\pi} d \sin^{3} \alpha}{4} \exp\left(-\frac{d^{2} \cos^{2} \alpha}{8\sigma^{2}}\right) \biggl[ 1 - \Phi\left(\frac{\sqrt{2} d \sin \alpha}{4\sigma}\right) \biggr] \biggl( \frac{\sqrt{\pi} \cos^{2} \theta}{2 \tan \alpha} - \sin 2\theta - \frac{\sin^{2} \theta}{\tan \alpha} \biggr)\)}  \\
		&-\scalebox{1.1}{\(\frac{\sqrt{2\pi} d \sin^{3} \beta}{4} \exp\left(-\frac{d^{2} \cos^{2} \beta}{8\sigma^{2}}\right) \biggl[ 1 - \Phi\left(\frac{\sqrt{2} d \sin \beta}{4\sigma}\right) \biggr] \biggl( \frac{\sqrt{\pi} \cos^{2} \theta}{2 \tan \beta} + \sin 2\theta - \frac{\sin^{2} \theta}{\tan \beta} \biggr)\)}  \\
		&+\scalebox{1.1}{\(\frac{\sqrt{2\pi} \sin \beta}{2} \exp\left(-\frac{d^{2} \cos^{2} \beta}{8\sigma^{2}}\right) \biggl[ 1 - \Phi\left(\frac{\sqrt{2} d \sin \beta}{4\sigma}\right) \biggr] \biggl( \frac{d}{2} \sin 2\theta + \frac{\sqrt{\pi} d \cos^{2} \theta}{4 \tan \beta} \biggr) \)}\\
		&+\scalebox{1.1}{\(\sigma\exp\left(-\frac{d^{2}}{8\sigma^{2}}\right)\left(\sin^{2}\alpha\biggl(\frac{\sqrt{\pi}\cos^{2}\theta}{2\tan\alpha}-\sin2\theta-\frac{\sin^{2}\theta}{\tan\alpha}\biggr)+\sin^{2}\beta\biggl(\frac{\sqrt{\pi}\cos^{2}\theta}{2\tan\beta}+\sin2\theta-\frac{\sin^{2}\theta}{\tan\beta}\biggr)\right)\biggr)\)}.
			\label{eq6}
		\end{aligned}
	\end{equation}
\end{figure*}

Taking L-type corner model as follows. Corner, which locates at \( u=0 \), can be represented as: 

	\begin{equation} 
			\begin{aligned}
				&f\left(x,y\right) =\begin{cases}T_1, &y\geq\tan(\frac{\pi}{2}-\alpha)x,x<0;\\ & y\geq0,0\leq x,\\ T_2,&\text{otherwise}.\end{cases}
				\label{eq11}
			\end{aligned}
\end{equation} 
		
\begin{figure}[!t]
			\centering
			\includegraphics[width=2.9in]{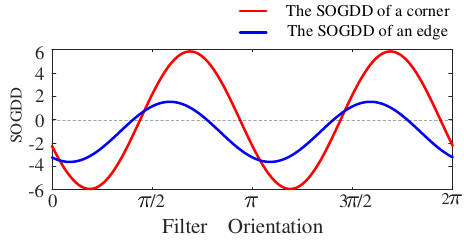}
			\caption{The END-type high-resolution corner model with the SOGDD representations of corner and edge point.}
			\label{fig_3}
\end{figure}
			
\begin{figure}[!t]
	\centering
	\includegraphics[width=2.9in]{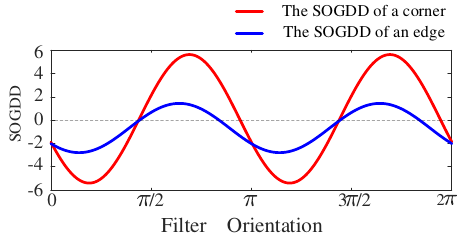}
	\caption{The L-type high-resolution corner model with the SOGDD representations of corner and edge point.}
	\label{fig_5}
\end{figure}
The SOGDD representation \(\Psi_{1}(\theta)\) of the corner, which locates at \( u=0 \),  please see equation (\ref{eq12}).

\begin{figure*}[b]
	\hrulefill
	\begin{equation}
		\begin{aligned}
			&\Psi_{1}(\theta)=\scalebox{1.1}{\(\int_{-\infty}^{+\infty}\int_{-\infty}^{+\infty} f(x,y)\psi(0-x,0-y) \,dx\,dy  =\frac{\left(T_{2}-T_{1}\right)}{2\pi\sigma^{2}}\bigg(-\sin2\theta +\sin^{2}\alpha\biggl(-\frac{\sqrt{\pi}\cos^{2}\theta}{2\tan\alpha}+\sin2\theta+\frac{\sin^{2}\theta}{\tan\alpha}\biggl)\bigg)\)}.
		\label{eq12}
		\end{aligned}
\end{equation}
\end{figure*}

Egde point, which locates at \( u=d/2 \), can be represented: 

		\begin{equation} 
			\begin{aligned}
				&f(x,y) = \begin{cases}
					T_1, &y \geq \tan\left(\frac{\pi}{2}-\alpha\right)(x + \frac{d}{2}), x < -\frac{d}{2};\\ & y \geq 0, -\frac{d}{2} \leq x, \\
					T_2, &\text{otherwise}.
				\end{cases}
				\label{eq13}
			\end{aligned}
		\end{equation} 
		
The SOGDD representation \(\Psi_{2}(\theta)\) of the edge point, which locates at \( u=d/2 \),  please see equation (\ref{eq14}).

\begin{figure*}[b]
	\begin{equation}
		\begin{aligned}
			\Psi_{2}(\theta) &= \scalebox{1.1}{\(\int_{-\infty}^{+\infty} \int_{-\infty}^{+\infty} f(x,y)\psi(0-x,0-y) \,dx\,dy\)}\\ 
			&=\scalebox{1.1}{\(\frac{\left(T_{2}-T_{1}\right)}{2\pi\sigma^{3}}\biggl(\sigma\sin^{2}\alpha \exp\left({-\frac{d^{2}}{8\sigma^{2}}}\right)\biggl(-\frac{\sqrt{\pi}\cos^{2}\theta}{2\tan\alpha}+\sin2\theta+\frac{\sin^{2}\theta}{\tan\alpha}\biggl)\)}\\
			&-\scalebox{1.1}{\(\sigma\sin2\theta\exp\biggl(-\frac{d^{2}}{8\sigma^{2}}\biggr)\)}
			-\scalebox{1}{\(\frac{\sqrt{2\pi}d\sin^{3}\alpha}{4}\exp\left({-\frac{d^{2}\cos^{2}\alpha}{8\sigma^{2}}}\right)\biggl[1-\Phi\biggl(\frac{\sqrt{2}d\sin\alpha}{4\sigma}\biggl)\biggl]\biggl(-\frac{\sqrt{\pi}\cos^{2}\theta}{2\tan\alpha}+\sin2\theta+\frac{\sin^{2}\theta}{\tan\alpha}\biggl)\)}\\
			&+\scalebox{1.1}{\(\frac{\sqrt{2\pi}\sin\alpha}{2}\exp\left({-\frac{d^2\cos^2\alpha}{8\sigma^2}}\right)\biggl[1-\Phi\biggl(\frac{\sqrt{2}d\sin\alpha}{4\sigma}\biggr)\biggr]\biggl(\frac{d}{2}\sin2\theta-\frac{\sqrt{\pi}d\cos^{2}\theta}{4\tan\alpha}\biggr)\biggr)\)}.
		\label{eq14}
		\end{aligned}
	\end{equation}
\end{figure*}

It is worth noting that the SOGDD representations of corner and edge point is difficult to determine the actual size, as shown in fig \ref{fig_3}. Only when the SOGDD representations at the corner is always greater than that at the edge point, can the corner be accurately detected. Therefore, only when the sum of the integrals of the square directional derivatives at the corner is greater than the sum of the integrals of the square derivatives at the edge point, can we accurately detect the corner, as shown below:
 
 \begin{equation} 
 	\begin{aligned}
 		\int_{0}^{2\pi} \Psi_{1}^{2}(\theta){} d\theta>\int_{0}^{2\pi} \Psi_{2}^{2}(\theta){} d\theta,
 		\label{eq15}
 	\end{aligned}
 \end{equation} 

 We further scale equation(\ref{eq15}) and ultimately derive the following quadratic equation. This equation indicates that the intrinsic physical properties of angles and edges are independent of grayscale values. At this point, the scaling factor \(\sigma \) is only related to the pixel value \(d \) and the angles \(\alpha \) and \(\beta \). Assuming that \(\sigma \) is a constant, further scaling of the equation produces a univariate quadratic inequality about \(\sigma \), where coefficients \(A \), \(B \), and \(C \) are all functions of pixel values \(d \) and constant values of angles \(\alpha \) and  \(\beta \). In the case of END-type corner model, We have obtained the following results:
 
\begin{equation} 
	\begin{aligned}
 	8A_{1}\sigma^{2}-4\sqrt{2\pi}dC_{1}\sigma-\pi d^{2}B_{1}>0,
 \label{eq16}
 \end{aligned}
\end{equation} 
 thus
 \begin{align}
 	&\sigma_{11}=\frac{\sqrt{2\pi}d(C_{1}-\sqrt{A_{1}B_{1}+C_{1}^{2}})}{4A_{1}},\notag \\
 	&\sigma_{21}=\frac{\sqrt{2\pi}d(C_{1}+\sqrt{A_{1}B_{1}+C_{1}^{2}})}{4A_{1}}.
 	 \label{eq17}
 \end{align}
 
 After verification, when the quadratic inequality has double roots, the opening is downward, and \(\sigma_{21} < \sigma_{11}\), \(d = 3\), The results obtained from the END-type high-resolution corner model are as follows: The minimum value of  \(\sigma_{21}\) is \(-25.307\) . The minimum value of  \(\sigma_{11}\) is \(1.204\).
 
 In terms of Equations (\ref{eq4}) and (\ref{eq6}), The END-type high-resolution corner model with the SOGDD representations of corner and edge point with we set
 \(T_1=50\), \(T_2=100\), \(\alpha=\pi/8\), \(\beta=\pi/3\) , \(d=3\) and \(\sigma=1.12 \), are show in fig \ref{fig_3}. Therefore, when the directional derivatives at the corner is larger than the derivatives at the edge point, can we accurately detect the corner.
  
The SOGDD representing corner and edge point is obtained through the L-type high-resolution corner model, we obtain the following equation from equation (\ref{eq15}). Our final derivation result is as follows:
 \begin{equation} 
 	\begin{aligned}
 		8A_{3}\sigma^{2}-4\sqrt{2\pi}dC_{3}\sigma-\pi d^{2}B_{3}>0,
 		\label{eq20}
 	\end{aligned}
 \end{equation} 
thus
 \begin{align}
 	&\sigma_{13}=\frac{\sqrt{2\pi}d(C_{3}-\sqrt{A_{3}B_{3}+C_{3}^{2}})}{4A_{3}},\notag \\
 	&\sigma_{23}=\frac{\sqrt{2\pi}d(C_{3}+\sqrt{A_{3}B_{3}+C_{3}^{2}})}{4A_{3}}.
 	\label{eq21}
 \end{align}

 After verification, the results obtained from the L-type high-resolution corner model are as follows: when the quadratic inequality has double roots, the opening is downward, and \(\sigma_{23} < \sigma_{13}\), \(d = 3\),  The minimum value of \(\sigma_{23}\) is \(-2.004\). The minimum value of \(\sigma_{13}\) is \(1.253\). 
  
 In terms of Equations (\ref{eq12}) and (\ref{eq14}), The L-type high-resolution corner model with the SOGDD representations of corner and edge point with we set \(T_1=50\), \(T_2=100\), \(\alpha=\pi/8\) , \(d=3\) and \(\sigma=1.15 \), are show in fig \ref{fig_5}.

 Through the derivation of Equation (\ref{eq15}), we obtained quadratic inequalities with different coefficients for three different high-resolution corner models, in summary, which leads to the result of Equation (\ref{eq15}). And it was proved in~\cite{zhang2021corner} that when \(\sigma^{2}<1\) is used, the variance of the image noise smoothed by the second-order generalized (including isotropic and anisotropic) Gaussian directional derivative (SOGGDD) filter will increase, thus \(\sigma^{2}>1\). So From Equation (\ref{eq15}) we have obtained for the first time the range of scale factors that has the ability to accurately depict high-resolution corners in the three high-resolution corner models, that is \(\sigma\in(1,1.2)\).
 
 \subsection{The Proposed Method}
 
 Images are 2D discrete signal images in the integer lattice \(\mathbb{Z}^2\). Using Equation (\ref{eq15}), we deduce that the value of \(\sigma=1.2\). Given \(K\) filter orientations \(\theta_{K}=\frac{\left(k-1\right)\pi}{K},(k=1,...,K)\), first discretize the continuous SOGDD filters:
 \begin{align}
 	\psi_{\sigma,k}(\mathbf{n})=\frac{1}{\sigma^{2}}\biggl(\frac{1}{\sigma^{2}}\bigl(\bigl[\cos\theta_{k}\sin\theta_{k}\bigr]\mathbf{n}\bigr)^{2}-1\biggr)g_{\sigma}\bigl(\mathbf{n}\bigr),
 	\label{eq22}
 \end{align}
 with 
 \[
 \mathbf{R}_{\mathbf{k}}=
 \begin{bmatrix}
 	\cos\theta_{k} & \sin\theta_{k} \\
 	-\sin\theta_{k} & \cos\theta_{k}
 \end{bmatrix},
 \quad
 \mathbf{n}=
 \begin{bmatrix}
 	n_{x} \\
 	n_{y}
 \end{bmatrix} \in \mathbb{Z}^{2}
 \]
 where \(\mathbf{n}\) represents the pixel coordinate within the integer lattice  \(\mathbb{Z}^2\). Obtain the discretized SOGDDs as:
 
 \begin{align}
 	\ell_{\iota\sigma.k}(\mathbf{n})=\sum_{m_{x}}\sum_{m_{y}}I(\mathbf{n}-\mathbf{m})\psi_{\sigma.k}(\mathbf{m}), \mathbf{m}=\left[m_{x},m_{y}\right]^{\tau}\in\mathbb{Z}^{2}.
 	\label{eq23}
 \end{align}
 
In the discrete domain, a new corner measurement is proposed as follows: Given a scale factor \(\sigma\), for each image pixel \(\left(n_{x},n_{y}\right)\), select an image block with \((p+1)\times(q+1)\) pixels centered on \(\left(n_{x},n_{y}\right)\).  In the image block, the absolute value of the SOGDDs in different filter orientations on each pixel, \(\varpi\left(n_{x},n_{y}\right)=\left[\ell_{1}\left(n_{x},n_{y}\right), \ell_{2}\left(n_{x},n_{y}\right), \cdots, \ell_{K}\left(n_{x},n_{y}\right)\right]\), is used to construct the matrix \( \Gamma_{K}\left(n_{x},n_{y}\right)=[\varpi^{T}(n_{x}-\frac{p}{2}\),\(n_{y}+\frac{q}{2}, \varpi^{T}\left(n_{x}-\frac{p}{2},n_{y}-\frac{q}{2}+1\right), \cdots, \varpi^{T}\left(n_{x}+\frac{p}{2},n_{y}+\frac{q}{2}\right)]\). \(\Gamma_{K}\left(n_{x},n_{y}\right)\) is a matrix with \(K\) rows and \((p+1)\times(q+1)\) columns. Then, matrix \(\Gamma_{K}\left(n_{x},n_{y}\right)\) is multiplied by its corresponding transpose matrix to obtain the \(K\times K\) autocorrelation matrix \(\Lambda_{K}\left(n_{x},n_{y}\right)\), also known as the second-order directional derivative correlation (SODDC) matrix.

\begin{figure}[!t]
	\centering
	\includegraphics[width=3.4in]{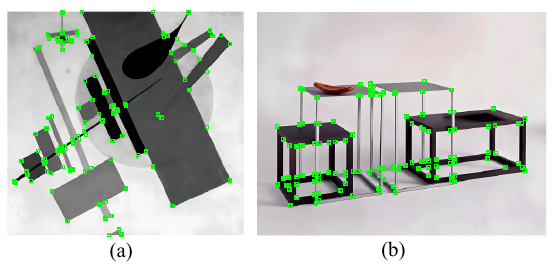}
	\caption{Test images (a) ‘Geometric’ and (b) ‘Table’ and their ground truths corner positions.}
	\label{fig_6}
\end{figure}

\begin{figure}[!t]
	\centering
	\includegraphics[width=3.5in]{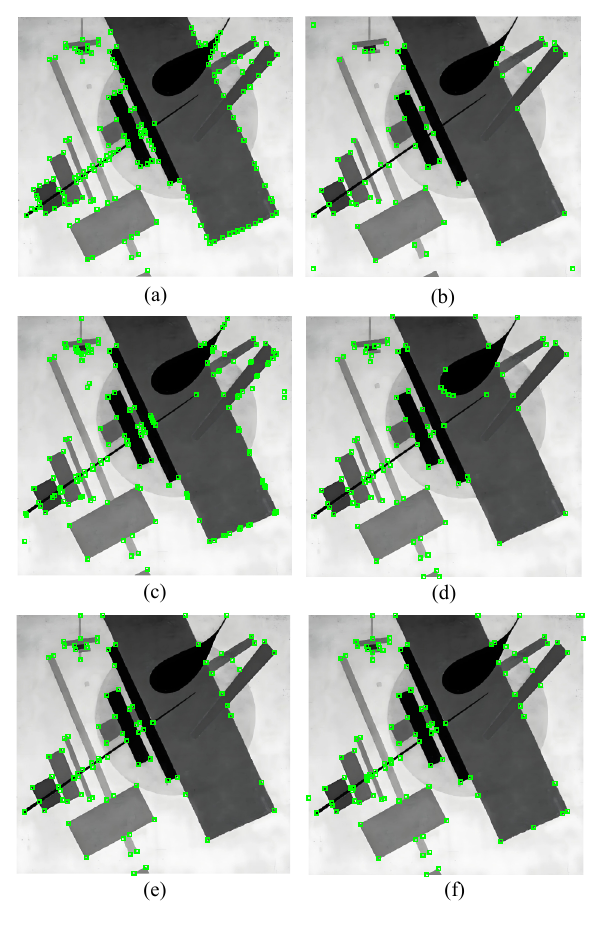}
	\caption{The results of corner detection on the test image ‘Geometric’. (a) Harris, (b) Harris-Laplace, (c) FSAT, (d) Zhang(2013), (e) Zhang(2021) (f) Proposed.}
	\label{fig_7}
\end{figure}

\begin{figure}[!t]
	\centering
	\includegraphics[width=3.5in]{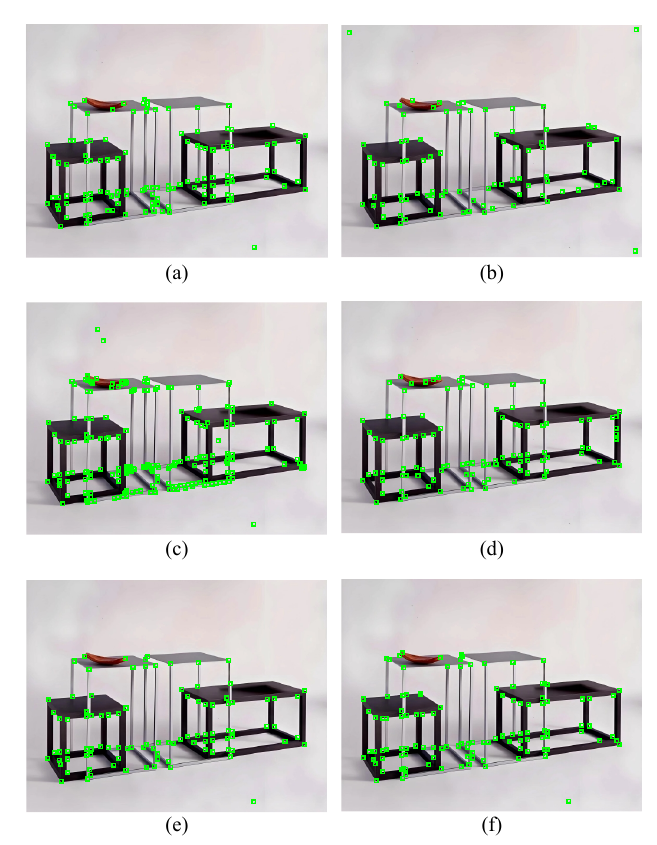}
	\caption{The results of corner detection on the test image ‘Table’. (a) Harris, (b) Harris-Laplace, (c) FSAT, (d) Zhang(2013), (e) Zhang(2021) (f) Proposed.}
	\label{fig_8}
\end{figure}
 
\begin{figure*}[hb]
	\hrulefill
	\begin{equation}
		\begin{aligned}
 		&\Lambda(n_{x},n_{y})={
 			\resizebox{0.9\linewidth}{!}{$
 				\begin{bmatrix}
 					\displaystyle\sum_{i=-\frac{p}{2}}^{\frac{p}{2}} \displaystyle\sum_{j=-\frac{q}{2}}^{\frac{q}{2}} \ell_{1}^{2}(n_{x}+i,n_{y}+j) & \cdots & \displaystyle\sum_{i=-\frac{p}{2}}^{\frac{p}{2}} \displaystyle\sum_{j=-\frac{q}{2}}^{\frac{q}{2}}\ell_{1}(n_{x}+i,n_{y}+j) \ell_{K}(n_{x}+i,n_{y}+j) \\
 					\vdots & \ddots & \vdots \\
 					\displaystyle\sum_{i=-\frac{p}{2}}^{\frac{p}{2}} \displaystyle\sum_{j=-\frac{q}{2}}^{\frac{q}{2}}\ell_{K}(n_{x}+i,n_{y}+j) \ell_{1}(n_{x}+i,n_{y}+j) & \cdots & \displaystyle\sum_{i=-\frac{p}{2}}^{\frac{p}{2}} \displaystyle\sum_{j=-\frac{q}{2}}^{\frac{q}{2}} \ell_{K}^{2}(n_{x}+i,n_{y}+j)  
 				\end{bmatrix}
 				$}}
 		\label{eq24}
		\end{aligned}
\end{equation}
\end{figure*}
 
The eigenvalues \(\{\lambda_1,\lambda_2,\ldots,\lambda_{K}\}\) of the matrix \(\Lambda_{K}\left(n_{x},n_{y}\right)\) are used to select candidate corners from the input image:

\begin{align}
	\Upsilon(n_x,n_y)=\frac{\prod_{k=1}^K\lambda_k}{\sum_{k=1}^K\lambda_k+\varsigma},
	\label{eq25}
\end{align}

where \(\varsigma\) is a small constant \(\left(\varsigma=2.22\times10^{-16}\right)\)
to avoid the occurrence of singular denominators. Equations  (\ref{eq25}) is used to calculate the corner measurement \(\Upsilon(n_x,n_y) \), which is the ratio of the determinant of matrix  \(\Lambda_{K}\left(n_{x},n_{y}\right)\) to the trace. Store the obtained corner metrics in the metric matrix, find the coordinates of all local maximum values and their corresponding maximum values, and determine whether the corresponding metric value for each coordinate exceeds the specified threshold to filter out effective corners.

\section{Experimental Results}
In this section, the proposed corner detector is compared with other state-of-the-art detectors in terms of detection accuracy, image matching quality as represented in~\cite{dusmanu2019d2}, robustness to afﬁne
transformations, noise in~\cite{awrangjeb2008robust} and 3D reconstruction quality as introduced in~\cite{xia2014accurate}.
We set the parameters  \(\sigma^{2}=1.2\), \((p+1)\times(q+1)=7\times7\) and \(\left(T_{h}=1\times10^{9}\right)\).

\subsection{Evaluation on Ground Truth Images}

Two commonly used images, `Geometry' and `Table', are used for evaluation accuracy evaluation. The basic facts of the two test images are shown in fig \ref{fig_6}. The image `Geometry' contains 114 corners, while the image `Table' contains 124 corners. The proposed corner detector is compared with the benchmark corner detectors in~\cite{mikolajczyk2004scale, rosten2008faster, zhang2021corner,harris1988combined,shui2013corner}. In this experiment, \(DC=\{(\hat{x}_{i},\hat{y}_{i}), i=1,2,\ldots,M_{1}\}\) represents the set of corners detected by a corner detector
and \(GT=\{(x_{j},y_{j}),j=1,,2,\ldots,M_{2}\}\) denotes a collection of real corners in ground truth images. For a corner\((x_{j},y_{j})\) in set \(GT\), we find the minimal distance is not more than a predefined threshold \(\delta\) here \( \delta=2\), corner \((x_{j},y_{j})\) is treated as correctly detected, and corner \((x_{j},y_{j})\) in set GT and the detected corner in set DC form a matched pair. Otherwise, the detected corner\((\hat{x}_{i},\hat{y}_{i})\)  in set DC is marked as a false corner. The localization error is defined as the average distance on all the matched pairs. 
Let \((\hat{x}_{l},\hat{y}_{l}),((x_{l},y_{l}): i=1,2,\ldots,N_{m}\) be the matched corner pairs in sets GT and DC. The localization error is calculated as the average distance for all the matched corner pairs. The average localization error is calculated by

\begin{align}
	L_e=\sqrt{\frac1{N_m}\sum_{l=1}^{N_m}(\left(\hat{x}_l-x_l\right)^2+\left(\hat{y}_l-y_l\right)^2)},
	\label{eq26}
\end{align}

where \({N_m}\) represents the number of matched corner pairs. The false corners, the missed corners, and the localization error are used to evaluate the performance for each method.

The GTs for the two test images are illustrated in fig \ref{fig_6} respectively. The threshold for each interest point detection method is properly tuned to achieve its best test results. The test results are shown in fig \ref{fig_7} and fig \ref{fig_8}. Table~\ref{tab:table1} lists the performance metrics of these six methods. From Table~\ref{tab:table1}, it can be seen that the proposed corner detector achieves high-resolution corner detection accuracy and minimal localization error.

\begin{table}[!t]
	\centering
	\caption{Comparison results of three corner detection methods with GT on two test images (``Geometric'' and "Table").\label{tab:table1}}
	\renewcommand{\arraystretch}{1.3}
	\begin{tabular}{ccccccc}
		\hline
		\textbf{}          & \multicolumn{2}{c}{Missed corners} & \multicolumn{2}{c}{False corners} & \multicolumn{2}{c}{Localization error} \\ \cline{2-7} 
		\textbf{Detectors}  & \makebox[0.7cm]{`Table'} & \makebox[0.7cm]{`Geometric'} & \makebox[0.7cm]{‘Table’} & \makebox[0.7cm]{‘Geometric’} & \makebox[0.7cm]{‘Table’} & \makebox[0.7cm]{‘Geometric’} \\ \hline
		Harris     & 19                  & 18            & 17                & 69             & 1.4409              & 1.3769              \\	
		Harris-Laplace     & 41                  & 49            & 69                & 57             & 1.6862             & 1.5038              \\		         
	FAST               & 12                  & 13            & 100               & 58             & 1.5469              & 1.5733            \\
	Zhang(2013)               & 35                  & 32            & 16                & 9             & 1.7769               & 1.6528           \\
	Zhang(2021)               & 35                  & 25            & 9                & 2             & 1.4649              & 1.1756            \\
	Proposed                    & 21                  & 17             & 21                & 10            & 1.4380              & 0.7995               \\ \hline
	\end{tabular}
\end{table}

\subsection{Average Repeatability Under IATs}

The average repeatability \(R_{\mathrm{avg}}\) is used to evaluate the robustness of interest point detection methods by comparing the matching of detected interest points between the reference image and the deformed image~\cite{awrangjeb2008robust}. 
Specifically, let \(L_{b}\) and \(L_{d}\) represent the number of detected interest points in the reference image and the deformed image, respectively. Denote \(F_{i}=\left(x_{fi},y_{fi}\right) \) as the interest points detected from the reference image and \(D_{j}=\left(x_{dj},y_{dj}\right)\) as the corresponding pixel positions in the deformed image. If the position \(D_{j}\) in the deformed image is within a predefined pixel distance (set to 4 pixels) from the position \(F_{i}\) in the reference image, it is considered a successful match, forming a matched interest point pair. The total number of matched interest point pairs is denoted by \(L_{r}\). The average repeatability \(R_{\mathrm{avg}}\) is obtained by calculating the ratio of the number of matched pairs \(L_{r}\) to the total number of detected interest points in the reference image and the deformed image \(L_{b}\) and \(L_{d}\).Then the average repeatability is 

\begin{align}
	R_{ang}=\frac{L_{r}}{2}\biggl(\frac{1}{L_{b}}+\frac{1}{L_{d}}\biggr)
	\label{eq27}
\end{align}

\begin{figure}[!t]
	\centering
	\includegraphics[width=3.5in]{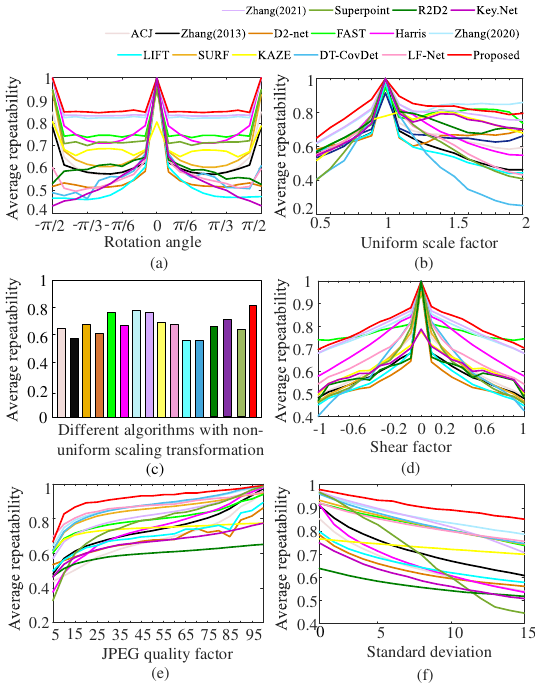}
	\caption{Average repeatability metrics for the fifteen methods under (a) rotation,(b) isotropic scaling, (c) anisotropic scaling, (d) shearing, (e) JPEG compression, (f) Gaussian noise.}
	\label{fig_9}
\end{figure}

The higher the average repeatability, the better the performance of the corresponding method. Evaluate the average repeatability of 16 methods in 30 images with various scenes but no real scenes. Apply six different sets of image geometric transformations on each image: (1) Rotation: Rotate the image using 18 different angles (excluding 0) within the range of \([-\pi/2,\pi/2]\). (2) Isotropic scaling: Uniformly scaling the image with scaling factors ranging from \([0.5, 2]\) and intervals of \(0.1\) (excluding 1.0). (3) Anisotropic scaling: The scaling factor range of the image in the horizontal direction is \([0.7, 1.5]\), and the scaling factor range in the vertical direction is \([0.5, 1.8]\), with an interval of \(0.1\). (4) Shearing: with a shear factor \(c\) by uniform sampling in \([1, 1]\) and intervals of \(0.1\) (excluding 0). (5) JPEG compression: JPEG compression of images,, with a quality factor range of \([5, 100]\) and an interval of \(5\). (6) Gaussian noise: Gaussian white noise is added to an image, with a mean of zero and a standard deviation range of \([1, 15]\). Using this evaluation criterion, the average repeatability measurement of 16 methods with default adjustable parameters was evaluated, by applying geometric transformations such as rotation, isotropic scaling, anisotropic scaling, shearing, JPEG compression, and Gaussian noise on each image, as shown in fig \ref{fig_9}. It can be seen that among the following 16 methods (FAST~\cite{rosten2008faster}, LF-Net~\cite{ono2018lf}, D2-Net~\cite{dusmanu2019d2}, LIFT~\cite{yi2016lift}, Harris~\cite{harris1988combined}, DT-CovDet~\cite{zhang2017learning},Zhang(2020)~\cite{zhang2020corner},Zhang(2021)~\cite{zhang2021corner}, ACJ~\cite{xia2014accurate}, Zhang(2013)~\cite{shui2013corner}, SURF~\cite{bay2006surf},  KAZE~\cite{alcantarilla2012kaze}, Superpoint~\cite{detone2018superpoint}, R2D2~\cite{revaud2019r2d2} and  Key.Net~\cite{barroso2019key}), our proposed high-resolution corner detection method has the best average repeatability.

\subsection{Image Matching}

\begin{figure}[!t]
	\centering
	\includegraphics[width=3.5in]{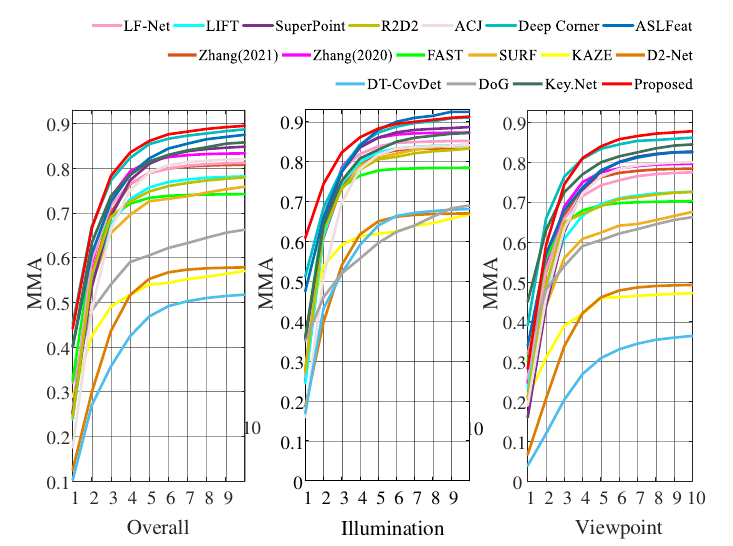}
\caption{Comparison results of the mean matching accuracy for the seventeen methods on the HPatches dataset.}
	\label{fig_10}
\end{figure}

In this experiment, the HPatches dataset~\cite{balntas2017hpatches} is employed to evaluate the mean matching accuracy (MMA)~\cite{dusmanu2019d2} for seventeen methods. Given two images \(F_{1}\) and \(F_{2}\) with corresponding homography matrix \( \mathbf{H} \), interest points detected in these images are denoted as \(\varpi\) and \(\varsigma\) , and their respective descriptors as  \(\iota_{1}\) and \(\iota_{2}\). When the nearest neighbors of descriptors in  \(\iota_{1}\) and \(\iota_{2}\) correspond to interest points in \(\varsigma\) and \(\varpi\), those interest points are considered a potential matching pair. The total number of potential matching pairs  \(\varsigma\) and \(\varpi\)  ( \(i = 1, 2, \ldots, N_{\text{possible}}\) ) is denoted as \(N_{\text{possible}}\). The distance error between the corresponding homogeneous coordinates of a pair \(\{\varsigma_i, \varpi_i\}\), represented as \((x_{\varsigma_i},y_{\varsigma_i},1) \text{ and } (u_{\varpi_i},v_{\varpi_i},1)\). So the distance error based on the known homography matrix can be calculated using the Euclidean norm:
\begin{align}
	&D_{\mathrm{error}} =||(x_{\varsigma_i},y_{\varsigma_i})^\top-(u_{\varpi_i}^{\prime\prime},v_{\varpi_i}^{\prime\prime})^\top||_2, \notag\\
	\text{with}\quad \quad 
	&(u_{\varpi_{i}}',v_{\varpi_{i}}',w_{\varpi_{i}}')^{\top}=H(u_{\varpi_{i}},v_{\varpi_{i}},1)^{\top}, \notag\\
	&(u_{\varpi_i}^{\prime\prime},v_{\varpi_i}^{\prime\prime})=(u_{\varpi_i}^{\prime}/w_{\varpi_i}^{\prime},v_{\varpi_i}^{\prime}/w_{\varpi_i}^{\prime}).
	\label{eq28}
\end{align}
﻿
If \(D_{\mathrm{error}}\) is less than a given pixel error threshold \(\left(P_{th},P_{th}\in\{1,2,\ldots,10\}\right)\) Possible matching pairs are marked as true matching pairs. The actual number of matches is defined as \(M_{\mathrm{match}}\). Then, the average matching accuracy between the two images is defined as:
﻿
\begin{align}
	\ MMA=\frac{N_{\mathrm{match}}}{N_{\mathrm{possible}}}.
	\label{eq29}
\end{align}
﻿
The HPatches dataset ~\cite{balntas2017hpatches} comprises 58 scenes for viewpoint transformations, 58 scenes for illumination changes, and each scene contains 6 images. We followed the D2-Net method~\cite{dusmanu2019d2}, selecting 108 out of 116 sequences and discarding \(8\) sequences with resolutions exceeding  \(1200\times1600\) pixels, as some methods were unable to process higher resolution images. In the HPatches image matching benchmark test, all methods (LIFT~\cite{yi2016lift}, LF-Net~\cite{ono2018lf}, SuperPoint~\cite{detone2018superpoint}, D2-Net~\cite{dusmanu2019d2}, DoG, Zhang (2020)~\cite{zhang2020corner}, Zhang (2021)~\cite{zhang2021corner}, DT-CovDet~\cite{zhang2017learning}, Key.Net~\cite{barroso2019key}, SURF~\cite{bay2006surf}, KAZE~\cite{alcantarilla2012kaze}, ACJ~\cite{xia2014accurate}, FAST~\cite{rosten2008faster}, Deep Corner~\cite{zhao2023deep}, ASLFeat~\cite{luo2020aslfeat}, and R2D2~\cite{revaud2019r2d2}) used default adjustable parameters and the results reported in their paper. Our method uses a 5-layer image pyramid, removes boundary pixels by 75 pixels, SOSNet as the descriptor, and uses the direction map of REKD~\cite{lee2022self} to filter the interest points. fig \ref{fig_10} shows the mean matching accuracy results, covering overall performance, illumination changes, and viewpoint changes. The matching threshold ranges from \(1\) to \(10\)  pixels, and the results are averaged across all image pairs. It can be seen from fig \ref{fig_10} that the proposed corner detector rank among the top in terms of performance.

\subsection{3D Reconstruction}

We conduct experiments on five datasets, (i.e., Fountain~\cite{schonberger2017comparative}, Herzjesu~\cite{schonberger2017comparative}, South Building~\cite{schonberger2016structure}, Madrid Metropolis~\cite{wilson2014robust} and Gendarmenmarkt~\cite{wilson2014robust}). to evaluate our method for the 3D reconstruction task and compared it with existing methods. Following the standards proposed in~\cite{schonberger2017comparative}, we first calibrated the cameras using structure-from-motion (SfM)~\cite{schonberger2016structure}, and then further obtained dense reconstruction of the given scene using multi-view stereo (MVS)~\cite{schonberger2016pixelwise} based on the output of SfM. In this experiment, we compared seven methods(i.e., DoG~\cite{lowe2004distinctive}, SuperPoint~\cite{detone2018superpoint}, D2-Net~\cite{dusmanu2019d2}, Zhang(2021)~\cite{zhang2021corner}, Zhang(2023)~\cite{zhang2023image} and the proposed corner  detection). To obtain comparable metrics, we did not employ the nearest neighbor ratio test mentioned in~\cite{schonberger2017comparative} and~\cite{tian2020d2d}, in order to better showcase the performance differences of each method. For the SfM models, the number of registered images, the number of reconstructed sparse points, the mean track lengths of the 3D points, and the mean reprojection error are summarized. For the MVS models, the number of reconstructed dense points are summarized. The results for the 3D reconstruction of the six methods are summarized in Table~\ref{tab:table2}. Our algorithm shows significant advantages in terms of the number of registered images, the number of reconstructed sparse points, and the extraction of dense points in five datasets. This indicates that the algorithm has high comprehensiveness and accuracy in feature point detection. Meanwhile, our algorithm also has advantages in the mean track lengths of the 3D
points, indicating that our algorithm tracks feature points stably and accurately, and can maintain consistency over relatively long paths, which plays a key role in improving the spatiotemporal continuity of point clouds.
Although our algorithm does not have the lowest mean reprojection error on certain datasets, it still maintains a low level of error on multiple datasets. In summary, our algorithm can ensure low mean reprojection error while maintaining high point density and long trajectory length. From Table 2, it can be seen that the proposed corner detection methods achieved the best and second best overall performance, respectively. This further confirms the superiority of our proposed algorithm.

\begin{table}[!t]
	\caption{Comparison results for SfM and MVS ($\dagger$ denotes that the result is provided by ~\cite{tian2020d2d}).\label{tab:table2}}
	\centering
	\renewcommand{\arraystretch}{1.3}
	\begin{tabular}{>{\centering\arraybackslash}p{0.8cm}>{\centering\arraybackslash}p{1cm}>{\centering\arraybackslash}p{1cm}>{\centering\arraybackslash}p{0.65cm} >{\centering\arraybackslash}p{0.65cm}>{\centering\arraybackslash}p{0.65cm}>{\centering\arraybackslash}p{0.65cm}}
		\toprule
		
		&        & \#Reg. & \#Sparse & \#Dense  & Track & Reproj. \\
		Dataset& Method & Images & Points &Points & Length & Error  \\
		&  &  &  &  &  & (pixels)  \\
		
		\midrule[0.85pt]
		
		Fountain & $\text{DoG}^{\dagger}$ & 11& 16K & 303K & 4.91 &{\bf 0.47}  \\
		11  & SupperPoint & 11& 15K & 305K & 4.58 &0.89  \\
		images & D2-Net & 11& 21K & 301K & 4.12 &1.49  \\
		& Zhang(2021) & 11& 28K & 305K & 4.97 &0.83 \\
		& Zhang(2023) & 11& 35K & 308K & 6.04 &0.67 \\
		& Proposed & 11&  35K &{\bf 309K} & {\bf6.15} &0.73  \\
		
		\midrule[0.85pt]
		
		Herzjesu & $\text{DoG}^{\dagger}$ & 8& 8K & 239K & 4.30 &{\bf 0.50}  \\
		8        & SuperPoint & 8& 5.6K & 242K & 4.41 &0.85  \\
		images   & D2-Net & 8& 15K & 224K & 3.16 &1.45  \\
		& Zhang(2021) & 8& 16K & 240K & 4.21 &0.82  \\
		& Zhang(2023) & 8& 18K & 245K & 4.52 &0.65  \\
		& Proposed & 8&  18K &{\bf 252K} & {\bf 4.61} &0.65  \\
		
		\midrule[0.85pt]
		
		South & $\text{DoG}^{\dagger}$ & 128& 159K & 2.12M & 5.18 &{\bf 0.62}  \\
		Building  & SuperPoint & 128& 127K & 2.13M & 7.17 &0.89  \\
		128 & D2-Net & 128& 183K & 2.11M & 3.55 &1.47 \\
		images    & Zhang(2021) & 128& 211K & 2.13M & 7.49 &0.84  \\
		& Zhang(2023) & 128& 223K & 2.19M & {\bf9.49 }&0.72  \\
		& Proposed & 128&  223K &{\bf 2.26M} & 7.73 &0.68  \\
		
		\midrule[0.85pt]
		
		Madrid & $\text{DoG}^{\dagger}$ & 793& 306K &  1.23M & 3.84&  0.93  \\
		Metro$-$  & SuperPoint & 715& 276K & 1.15M & 4.61 &1.24  \\
		polis & D2-Net & 795& 291K & 0.98M & 5.19 &1.41 \\
		1344  & Zhang(2021) & 802& 314K & 1.15M & 5.48 &  1.22  \\
	   images  & Zhang(2023) & 829&{\bf 372K} &{\bf 1.35M} &{\bf 5.97} &  1.11 \\
		& Proposed &{\bf 873}&  316K &1.31M &  5.79 & {\bf0.70}  \\  
		
		\midrule[0.85pt]
		Gendar$-$ & $\text{DoG}^{\dagger}$& 1018& 827K & 2.06M & 2.56 & 1.09 \\
		men$-$  & SuperPoint & 1129& 816K & 2.09M & 3.22 & 1.29 \\
		markt & D2-Net & 1227& 811K & 2.52M & 4.18 &1.48  \\
		1463    & Zhang(2021) & 1186& 820K & 3.03M & {\bf5.42} &1.21  \\
		images& Zhang(2023) & 1242& 855K & 3.40M & 5.31 &1.15  \\
		 & Proposed & {\bf 1263}&  {\bf 863K} &{\bf 3.48M} &  5.24 &   {\bf1.05}  \\
		\bottomrule
	\end{tabular}
\end{table}




\section{Conclusion}

In this work, the SOGDD representations of two type corners are derived which help us to demonstrate how to select a Gaussian filtering scale for accurately obtaining intensity variation information from images for depicting the high-resolution corners. Then, for the first time, a new image high-resolution corner detection method is proposed which has the capability to accurately detect adjacent corners. Furthermore, experimental results verify the superiority of the proposed method in terms of localization error, image matching,  robustness to afﬁne transformations, noise and 3D reconstruction.

\section*{Acknowledgments}
This should be a simple paragraph before the References to thank those individuals and institutions who have supported your work on this article.

%
\bibliographystyle{IEEEtran}
\bibliography{IEEEtran}

@article{lowe2004distinctive,
  title={Distinctive image features from scale-invariant keypoints},
  author={Lowe , David G},
  journal={International Journal of Computer Vision},
  pages={91--110},
  year={2004}
}

@inproceedings{luo2020aslfeat,
  title={{ASLFeat: Learning local features of accurate shape and localization}},
  author={Luo, Zixin and Zhou, Lei and Bai, Xuyang and Chen, Hongkai and Zhang, Jiahui and Yao, Yao and Li, Shiwei and Fang, Tian and Quan, Long},
  booktitle={Proceedings of the IEEE Conference on Computer Vision and Pattern Recognition},
  pages={6589--6598},
  year={2020}
}

@article{jing2022recent,
  title={Recent advances on image edge detection: A comprehensive review},
  author={Jing, Junfeng and Liu, Shenjuan and Wang, Gang and Zhang, Weichuan and Sun, Changming},
  journal={Neurocomputing},
  volume={503},
  pages={259--271},
  year={2022},
  publisher={Elsevier}
}

@article{wang2024unbiased,
  title={An unbiased feature estimation network for few-shot fine-grained image classification},
  author={Wang, Jiale and Lu, Jin and Yang, Junpo and Wang, Meijia and Zhang, Weichuan},
  journal={Sensors},
  volume={24},
  number={23},
  pages={7737},
  year={2024},
  publisher={MDPI}
}

@article{lei2024semi,
  title={Semi-Supervised 3-D Medical Image Segmentation Using Multiconsistency Learning With Fuzzy Perception-Guided Target Selection},
  author={Lei, Tao and Song, Wenbiao and Zhang, Weichuan and Du, Xiaogang and Li, Chenxia and He, Lifeng and Nandi, Asoke K},
  journal={IEEE Transactions on Radiation and Plasma Medical Sciences},
  volume={9},
  number={4},
  pages={421--432},
  year={2024},
  publisher={IEEE}
}

@article{an2023edge,
  title={Edge detection using multi-directional anisotropic Gaussian directional derivative},
  author={An, Ying and Jing, Junfeng and Zhang, Weichuan},
  journal={Signal, Image and Video Processing},
  volume={17},
  number={7},
  pages={3767--3774},
  year={2023},
  publisher={Springer}
}

@article{ren4962361adaptive,
  title={Adaptive feature selection-based feature reconstruction network for few-shot learning},
  author={Ren, Jie and An, Yaohui and Lei, Tao and Yang, Junpo and Zhang, Wenyue and Pan, Zicheng and Liao, Yi and Gao, Yongsheng and Sun, Changming and Zhang, Weichuan},
  journal={Pattern Recognition},
  year={2025}
}

@article{lu2023track,
  title={Track-before-detect algorithm based on cost-reference particle filter bank for weak target detection},
  author={Lu, Jin and Peng, Guojie and Zhang, Weichuan and Sun, Changming},
  journal={IEEE Access},
  volume={11},
  pages={121688--121701},
  year={2023},
  publisher={IEEE}
}

@article{zheng2023fully,
  title={Fully unsupervised domain-agnostic image retrieval},
  author={Zheng, Ziqiang and Ren, Hao and Wu, Yang and Zhang, Weichuan and Lu, Hong and Yang, Yang and Shen, Heng Tao},
  journal={IEEE Transactions on Circuits and Systems for Video Technology},
  volume={34},
  number={6},
  pages={5077--5090},
  year={2023},
  publisher={IEEE}
}

@article{wang2025principal,
  title={A principal component analysis-based feature optimization network for few-shot fine-grained image classification},
  author={Wang, Meijia and Zheng, Boyuan and Wang, Guochao and Yang, Junpo and Lu, Jin and Zhang, Weichuan},
  journal={Mathematics},
  volume={13},
  number={7},
  pages={1098},
  year={2025},
  publisher={MDPI}
}

@inproceedings{liao2022asrsnet,
  title={ASRSNet: Automatic Salient Region Selection Network for Few-Shot Fine-Grained Image Classification},
  author={Liao, Yi and Zhang, Weichuan and Gao, Yongsheng and Sun, Changming and Yu, Xiaohan},
  booktitle={International Conference on Pattern Recognition and Artificial Intelligence},
  pages={627--638},
  year={2022},
  organization={Springer}
}

@article{bao2022corner,
  title={A corner detection method based on adaptive multi-directional anisotropic diffusion},
  author={Bao, Junmin and Jing, Junfeng and Zhang, Weichuan and Liu, Chao and Gao, Tian},
  journal={Multimedia Tools and Applications},
  volume={81},
  number={20},
  pages={28729--28754},
  year={2022},
  publisher={Springer}
}

@article{lu2022image,
  title={Image local structure information learning for fine-grained visual classification},
  author={Lu, Jin and Zhang, Weichuan and Zhao, Yali and Sun, Changming},
  journal={Scientific Reports},
  volume={12},
  number={1},
  pages={19205},
  year={2022},
  publisher={Nature Publishing Group UK London}
}

@inproceedings{jing2021novel,
  title={A novel decision mechanism for image edge detection},
  author={Jing, Junfeng and Liu, Shenjuan and Liu, Chao and Gao, Tian and Zhang, Weichuan and Sun, Changming},
  booktitle={International Conference on Intelligent Computing},
  pages={274--287},
  year={2021},
  organization={Springer}
}

@article{islam2023background,
  title={Background-aware band selection for object tracking in hyperspectral videos},
  author={Islam, Mohammad Aminul and Zhou, Jun and Zhang, Weichuan and Gao, Yongsheng},
  journal={IEEE Geoscience and Remote Sensing Letters},
  volume={20},
  pages={1--5},
  year={2023},
  publisher={IEEE}
}

@article{gao2020fast,
  title={Fast corner detection using approximate form of second-order Gaussian directional derivative},
  author={Gao, Tian and Jing, Junfeng and Liu, Chao and Zhang, Weichuan and Gao, Yongsheng and Sun, Changming},
  journal={IEEE Access},
  volume={8},
  pages={194092--194104},
  year={2020},
  publisher={IEEE}
}

@article{li2023mutual,
  title={Mutual interference mitigation of millimeter-wave radar based on variational mode decomposition and signal reconstruction},
  author={Li, Yanbing and Feng, Bo and Zhang, Weichuan},
  journal={Remote Sensing},
  volume={15},
  number={3},
  pages={557},
  year={2023},
  publisher={MDPI}
}

@article{zhang2021ndpnet,
  title={NDPNet: A novel non-linear data projection network for few-shot fine-grained image classification},
  author={Zhang, Weichuan and Liu, Xuefang and Xue, Zhe and Gao, Yongsheng and Sun, Changming},
  journal={arXiv preprint arXiv:2106.06988},
  year={2021}
}

@article{ren2024few,
  title={Few-shot fine-grained image classification: A comprehensive review},
  author={Ren, Jie and Li, Changmiao and An, Yaohui and Zhang, Weichuan and Sun, Changming},
  journal={AI},
  volume={5},
  number={1},
  pages={405--425},
  year={2024},
  publisher={MDPI}
}

@article{liu2024aekan,
  title={AEKAN: Exploring Superpixel-based AutoEncoder Kolmogorov-Arnold Network for Unsupervised Multimodal Change Detection},
  author={Liu, Tongfei and Xu, Jianjian and Lei, Tao and Wang, Yingbo and Du, Xiaogang and Zhang, Weichuan and Lv, Zhiyong and Gong, Maoguo},
  journal={IEEE Transactions on Geoscience and Remote Sensing},
  year={2024},
  publisher={IEEE}
}

@article{li2019multi,
  title={Multi-scale anisotropic Gaussian kernels for image edge detection},
  author={Li, Yunhong and Bi, Yuandong and Zhang, Weichuan and Sun, Changming},
  journal={IEEE Access},
  volume={8},
  pages={1803--1812},
  year={2019},
  publisher={IEEE}
}

@inproceedings{wang2018survey,
  title={A survey of corner detection methods},
  author={Wang, Junqing and Zhang, Weichuan},
  booktitle={2018 2nd International Conference on Electrical Engineering and Automation (ICEEA 2018)},
  pages={214--219},
  year={2018},
  organization={Atlantis Press}
}

@article{jing2023ecfrnet,
  title={ECFRNet: Effective corner feature representations network for image corner detection},
  author={Jing, Junfeng and Liu, Chao and Zhang, Weichuan and Gao, Yongsheng and Sun, Changming},
  journal={Expert Systems with Applications},
  volume={211},
  pages={118673},
  year={2023},
  publisher={Elsevier}
}

@article{wang2020corner,
  title={Corner detection based on shearlet transform and multi-directional structure tensor},
  author={Wang, Mingzhe and Zhang, Weichuan and Sun, Changming and Sowmya, Arcot},
  journal={Pattern Recognition},
  volume={103},
  pages={107299},
  year={2020},
  publisher={Elsevier}
}

@article{qiu2021recurrent,
  title={Recurrent convolutional neural networks for 3D mandible segmentation in computed tomography},
  author={Qiu, Bingjiang and Guo, Jiapan and Kraeima, Joep and Glas, Haye Hendrik and Zhang, Weichuan and Borra, Ronald JH and Witjes, Max Johannes Hendrikus and van Ooijen, Peter MA},
  journal={Journal of personalized medicine},
  volume={11},
  number={6},
  pages={492},
  year={2021},
  publisher={MDPI}
}

@article{li2023traffic,
  title={Traffic flow digital twin generation for highway scenario based on radar-camera paired fusion},
  author={Li, Yanbing and Zhang, Weichuan},
  journal={Scientific reports},
  volume={13},
  number={1},
  pages={642},
  year={2023},
  publisher={Nature Publishing Group UK London}
}

@article{zhang2019corner,
  title={Corner detection using second-order generalized Gaussian directional derivative representations},
  author={Zhang, Weichuan and Sun, Changming},
  journal={IEEE transactions on pattern analysis and machine intelligence},
  volume={43},
  number={4},
  pages={1213--1224},
  year={2019},
  publisher={IEEE}
}

@article{zhang2014corner,
  title={Corner detection using Gabor filters},
  author={Zhang, Wei-Chuan and Wang, Fu-Ping and Zhu, Lei and Zhou, Zuo-Feng},
  journal={IET Image Processing},
  volume={8},
  number={11},
  pages={639--646},
  year={2014},
  publisher={Wiley Online Library}
}

@article{zhang2015contour,
  title={Contour-based corner detection via angle difference of principal directions of anisotropic Gaussian directional derivatives},
  author={Zhang, Wei-Chuan and Shui, Peng-Lang},
  journal={Pattern Recognition},
  volume={48},
  number={9},
  pages={2785--2797},
  year={2015},
  publisher={Elsevier}
}

@article{zhang2017noise,
  title={Noise robust image edge detection based upon the automatic anisotropic Gaussian kernels},
  author={Zhang, WeiChuan and Zhao, YaLi and Breckon, Toby P and Chen, Long},
  journal={Pattern Recognition},
  volume={63},
  pages={193--205},
  year={2017},
  publisher={Elsevier}
}

@article{shui2012noise,
  title={Noise-robust edge detector combining isotropic and anisotropic Gaussian kernels},
  author={Shui, Peng-Lang and Zhang, Wei-Chuan},
  journal={Pattern Recognition},
  volume={45},
  number={2},
  pages={806--820},
  year={2012},
  publisher={Elsevier}
}

@ARTICLE{6507646,
  author={Shui, Peng-Lang and Zhang, Wei-Chuan},
  journal={IEEE Transactions on Image Processing}, 
  title={Corner Detection and Classification Using Anisotropic Directional Derivative Representations}, 
  year={2013},
  volume={22},
  number={8},
  pages={3204-3218},
  doi={10.1109/TIP.2013.2259834}}

@article{pan2024pseudo,
  title={Pseudo-set frequency refinement architecture for fine-grained few-shot class-incremental learning},
  author={Pan, Zicheng and Zhang, Weichuan and Yu, Xiaohan and Zhang, Miaohua and Gao, Yongsheng},
  journal={Pattern Recognition},
  volume = {155},
  pages={110686},
  year={2024}
}

@article{zhang2024re,
  title={Re-abstraction and perturbing support pair network for few-shot fine-grained image classification},
  author={Zhang, Weichuan and Zhao, Yali and Gao, Yongsheng and Sun, Changming},
  journal={Pattern Recognition},
  volume={148},
  pages={110158},
  year={2024}
}

@article{liao2025dynamic,
  title={Dynamic accumulated attention map for interpreting evolution of decision-making in vision transformer},
  author={Liao, Yi and Gao, Yongsheng and Zhang, Weichuan},
  journal={Pattern Recognition},
  volume={165},
  pages={111607},
  year={2025}
}

@article{jing2022image,
  title={{Image feature information extraction for interest point detection: A comprehensive review}},
  author={Jing, Junfeng and Gao, Tian and Zhang, Weichuan and Gao, Yongsheng and Sun, Changming},
  journal={IEEE Transactions on Pattern Analysis and Machine Intelligence},
  volume={45},
  number={4},
  pages={4694--4712},
  year={2022}
}

@inproceedings{law2018cornernet,
  title={{CornerNet: Detecting objects as paired keypoints}},
  author={Law, Hei and Deng, Jia},
  booktitle={Proceedings of the European Conference on Computer Vision},
  pages={734--750},
  year={2018}
}

@inproceedings{mair2010adaptive,
  title={Adaptive and generic corner detection based on the accelerated segment test},
  author={Mair, Elmar and Hager, Gregory D and Burschka, Darius and Suppa, Michael and Hirzinger, Gerhard},
  booktitle={European Conference on Computer Vision},
  pages={183--196},
  year={2010}
}

@inproceedings{lucas1981iterative,
  title={An iterative image registration technique with an application to stereo vision},
  author={Lucas, Bruce D and Kanade, Takeo},
  booktitle={International Joint Conference on Artificial Intelligence},
  pages={674--679},
  year={1981}
}

@article{zhao2023deep,
  title={Deep corner},
  author={Zhao, Shanshan and Gong, Mingming and Zhao, Haimei and Zhang, Jing and Tao, Dacheng},
  journal={International Journal of Computer Vision},
  pages={2908--2932},
  year={2023}
}

@inproceedings{trujillo2006synthesis,
  title={Synthesis of interest point detectors through genetic programming},
  author={Trujillo, Leonardo and Olague, Gustavo},
  booktitle={Proceedings of the Conference on Genetic and Evolutionary Computation},
  pages={887--894},
  year={2006}
}

@inproceedings{carion2020end,
  title={End-to-end object detection with transformers},
  author={Carion, Nicolas and Massa, Francisco and Synnaeve, Gabriel and Usunier, Nicolas and Kirillov, Alexander and Zagoruyko, Sergey},
  booktitle={European Conference on Computer Vision},
  pages={213--229},
  year={2020}
}

@article{zhang2021corner,
  title={{Corner detection using second-order generalized Gaussian directional derivative representations}},
  author={Zhang, Weichuan and Sun, Changming},
  journal={IEEE Transactions on Pattern Analysis and Machine Intelligence},
  pages={1213--1224},
  year={2021}
}

@article{zhang2020corner,
  title={Corner detection using multi-directional structure tensor with multiple scales},
  author={Zhang, Weichuan and Sun, Changming},
  journal={International Journal of Computer Vision},
  pages={438--459},
  year={2020}
}

@article{zhang2023image,
  title={Image intensity variation information for interest point detection},
  author={Zhang, Weichuan and Sun, Changming and Gao, Yongsheng},
  journal={IEEE Transactions on Pattern Analysis and Machine Intelligence},
  pages={9883--9894},
  year={2023}
}

@article{zhang2019discrete,
  title={Discrete curvature representations for noise robust image corner detection},
  author={Zhang, Weichuan and Sun, Changming and Breckon, Toby and Alshammari, Naif},
  journal={IEEE Transactions on Image Processing},
  pages={4444--4459},
  year={2019}
}

@inproceedings{harris1988combined,
  title={A combined corner and edge detector},
  author={Harris, Chris and Stephens, Mike},
  booktitle={Alvey Vision Conference},
  pages={10--5244},
  year={1988}
}

@article{vaswani2017,
  title={Attention is all you need},
  author={Vaswani, Ashish and Shazeer, Noam and Parmar, Niki and Uszkoreit, Jakob and Jones, Llion and Aidan N. Gomez and Kaiser, Lukasz and Polosukhin, Illia },
  booktitle={Advances in Neural Information Processing Systems},
  volume={3},
  year={2017}
}

@article{rosten2008faster,
  title={{Faster and better: A machine learning approach to corner detection}},
  author={Rosten, Edward and Porter, Reid and Drummond, Tom},
  journal={IEEE Transactions on Pattern Analysis and Machine Intelligence},
  pages={105--119},
  year={2008}
}

@article{smith1997susan,
  title={{SUSAN: A new approach to low level image processing}},
  author={Smith, Stephen M and Brady, J Michael},
  journal={International Journal of Computer Vision},
  pages={45--78},
  year={1997}
}

@article{bay2006surf,
  title={{SURF: Speeded up robust features}},
  author={Bay, Herbert},
  booktitle={European Conference on Computer Vision},
  pages={346--359},
  year={2006}
}

@inproceedings{liu2016ssd,
  title={{SSD: Single shot multibox detector}},
  author={Liu, Wei and Anguelov, Dragomir and Erhan, Dumitru and Szegedy, Christian and Reed, Scott and Fu, Cheng-Yang and Berg, Alexander C},
  booktitle={European Conference on Computer Vision},
  pages={21--37},
  year={2016}
}

@inproceedings{detone2018superpoint,
  title={{Superpoint: Self-supervised interest point detection and description}},
  author={DeTone, Daniel and Malisiewicz, Tomasz and Rabinovich, Andrew},
  booktitle={Proceedings of the IEEE Conference on Computer Vision and Pattern Recognition Workshops},
  pages={224--236},
  year={2018}
}

@inproceedings{dusmanu2019d2,
  title={{D2-Net: A trainable CNN for joint description and detection of local features}},
  author={Dusmanu, Mihai and Rocco, Ignacio and Pajdla, Tomas and Pollefeys, Marc and Sivic, Josef and Torii, Akihiko and Sattler, Torsten},
  booktitle={Proceedings of the IEEE Conference on Computer Vision and Pattern Recognition},
  pages={8092--8101},
  year={2019}
}

@inproceedings{sattler2018benchmarking,
  title={{Benchmarking 6DOF outdoor visual localization in changing conditions}},
  author={Sattler, Torsten and Maddern, Will and Toft, Carl and Torii, Akihiko and Hammarstrand, Lars and Stenborg, Erik and Safari, Daniel and Okutomi, Masatoshi and Pollefeys, Marc and Sivic, Josef and others},
  booktitle={Proceedings of the IEEE Conference on Computer Vision and Pattern Recognition},
  pages={8601--8610},
  year={2018}
}

@inproceedings{taira2018inloc,
  title={{InLoc: Indoor visual localization with dense matching and view synthesis}},
  author={Taira, Hajime and Okutomi, Masatoshi and Sattler, Torsten and Cimpoi, Mircea and Pollefeys, Marc and Sivic, Josef and Pajdla, Tomas and Torii, Akihiko},
  booktitle={Proceedings of the IEEE Conference on Computer Vision and Pattern Recognition},
  pages={7199--7209},
  year={2018}
}

@article{revaud2019r2d2,
  title={{R2D2: Reliable and repeatable detector and descriptor}},
  author={Revaud, Jerome and De Souza, Cesar and Humenberger, Martin and Weinzaepfel, Philippe},
  booktitle={Advances in Neural Information Processing Systems},
  year={2019}
}

@inproceedings{kanakis2023zippypoint,
  title={{ZippyPoint: Fast interest point detection, description, and matching through mixed precision discretization}},
  author={Kanakis, Menelaos and Maurer, Simon and Spallanzani, Matteo and Chhatkuli, Ajad and Van Gool, Luc},
  booktitle={Proceedings of the IEEE Conference on Computer Vision and Pattern Recognition},
  pages={6114--6123},
  year={2023}
}

@inproceedings{edstedt2024dedode,
  title={{DeDoDe: Detect, don’t describe-describe, don’t detect for local feature matching}},
  author={Edstedt, Johan and B{\"o}kman, Georg and Wadenb{\"a}ck, M{\aa}rten and Felsberg, Michael},
  booktitle={International Conference on 3D Vision},
  pages={148--157},
  year={2024}
}

@inproceedings{zhao2024balf,
  title={{BALF: Simple and efficient blur aware local feature detector}},
  author={Zhao, Zhenjun},
  booktitle={Proceedings of the IEEE Winter Conference on Applications of Computer Vision},
  pages={3362--3372},
  year={2024}
}

@article{mikolajczyk2004scale,
  title={Scale \& affine invariant interest point detectors},
  author={Mikolajczyk, Krystian and Schmid, Cordelia},
  journal={International Journal of Computer Vision},
  pages={63--86},
  year={2004}
}

@article{ono2018lf,
  title={{LF-Net: Learning local features from images}},
  author={Ono, Yuki and Trulls, Eduard and Fua, Pascal and Yi, Kwang Moo},
  booktitle={Advances in Neural Information Processing Systems},
  year={2018}
}

@inproceedings{yi2016lift,
  title={{LIFT: Learned invariant feature transform}},
  author={Yi, Kwang Moo and Trulls, Eduard and Lepetit, Vincent and Fua, Pascal},
  booktitle={European Conference on Computer Vision},
  pages={467--483},
  year={2016}
}

@ARTICLE{barroso2019key,
  author={Barroso-Laguna, Axel and Mikolajczyk, Krystian},
  journal={IEEE Transactions on Pattern Analysis and Machine Intelligence}, 
  title={{Key.Net: Keypoint detection by handcrafted and learned CNN filters revisited}}, 
  year={2023},
  volume={45},
  number={1},
  pages={698-711}
}

@inproceedings{zhang2017learning,
  title={Learning discriminative and transformation covariant local feature detectors},
  author={Zhang, Xu and Yu, Felix X and Karaman, Svebor and Chang, Shih-Fu},
  booktitle={Proceedings of the IEEE Conference on Computer Vision and Pattern Recognition},
  pages={6818--6826},
  year={2017}
}

@article{xia2014accurate,
  title={Accurate junction detection and characterization in natural images},
  author={Xia, Gui-Song and Delon, Julie and Gousseau, Yann},
  journal={International Journal of Computer Vision},
  pages={31--56},
  year={2014}
}

@article{shui2013corner,
  title={Corner detection and classification using anisotropic directional derivative representations},
  author={Shui, Peng-Lang and Zhang, Wei-Chuan},
  journal={IEEE Transactions on Image Processing},
  pages={3204--3218},
  year={2013}
}

@inproceedings{alcantarilla2012kaze,
  title={{KAZE features}},
  author={Alcantarilla, Pablo Fern{\'a}ndez and Bartoli, Adrien and Davison, Andrew J},
  booktitle={European Conference on Computer Vision},
  pages={214--227},
  year={2012},
}

@article{awrangjeb2008robust,
  title={Robust image corner detection based on the chord-to-point distance accumulation technique},
  author={Awrangjeb, Mohammad and Lu, Guojun},
  journal={IEEE Transactions on Multimedia},
  pages={1059--1072},
  year={2008}
}

@inproceedings{balntas2017hpatches,
  title={HPatches: A benchmark and evaluation of handcrafted and learned local descriptors},
  author={Balntas, Vassileios and Lenc, Karel and Vedaldi, Andrea and Mikolajczyk, Krystian},
  booktitle={Proceedings of the IEEE Conference on Computer Vision and Pattern Recognition},
  pages={5173--5182},
  year={2017}
}

@inproceedings{lee2022self,
  title={Self-supervised equivariant learning for oriented keypoint detection},
  author={Lee, Jongmin and Kim, Byungjin and Cho, Minsu},
  booktitle={Proceedings of the IEEE Conference on Computer Vision and Pattern Recognition},
  pages={4847--4857},
  year={2022}
}

@inproceedings{schonberger2016structure,
  title={Structure-from-motion revisited},
  author={Schonberger, Johannes L and Frahm, Jan-Michael},
  booktitle={Proceedings of the IEEE Conference on Computer Vision and Pattern Recognition},
  pages={4104--4113},
  year={2016}
}

@inproceedings{schonberger2017comparative,
  title={Comparative evaluation of hand-crafted and learned local features},
  author={Schonberger, Johannes L and Hardmeier, Hans and Sattler, Torsten and Pollefeys, Marc},
  booktitle={Proceedings of the IEEE Conference on Computer Vision and Pattern Recognition},
  pages={1482--1491},
  year={2017}
}

@inproceedings{schonberger2016pixelwise,
  title={Pixelwise view selection for unstructured multi-view stereo},
  author={Sch{\"o}nberger, Johannes L and Zheng, Enliang and Frahm, Jan-Michael and Pollefeys, Marc},
  booktitle={European Conference on Computer Vision},
  pages={501--518},
  year={2016}
}

@inproceedings{tian2020d2d,
  title={{D2D: Keypoint extraction with describe to detect approach}},
  author={Tian, Yurun and Balntas, Vassileios and Ng, Tony and Barroso-Laguna, Axel and Demiris, Yiannis and Mikolajczyk, Krystian},
  booktitle={Proceedings of the Asian Conference on Computer Vision},
  year={2020}
}

@inproceedings{wilson2014robust,
  title={{Robust global translations with 1DSfM}},
  author={Wilson, Kyle and Snavely, Noah},
  booktitle={European Conference on Computer Vision},
  pages={61--75},
  year={2014}
}

@article{gao2023dynamic,
  title={Dynamic keypoint detection network for image matching},
  author={Gao, Yuan and He, Jianfeng and Zhang, Tianzhu and Zhang, Zhe and Zhang, Yongdong},
  journal={IEEE Transactions on Pattern Analysis and Machine Intelligence},
  year={2023}
}

@inproceedings{potje2024xfeat,
  title={XFeat: Accelerated Features for Lightweight Image Matching},
  author={Potje, Guilherme and Cadar, Felipe and Araujo, Andr{\'e} and Martins, Renato and Nascimento, Erickson R},
  booktitle={Proceedings of the IEEE Conference on Computer Vision and Pattern Recognition},
  pages={2682--2691},
  year={2024}
}

@article{zhao2022alike,
  title={ALIKE: Accurate and lightweight keypoint detection and descriptor extraction},
  author={Zhao, Xiaoming and Wu, Xingming and Miao, Jinyu and Chen, Weihai and Chen, Peter CY and Li, Zhengguo},
  journal={IEEE Transactions on Multimedia},
  pages={3101--3112},
  year={2022}
}

@article{zhao2023aliked,
  title={ALIKED: A lighter keypoint and descriptor extraction network via deformable transformation},
  author={Zhao, Xiaoming and Wu, Xingming and Chen, Weihai and Chen, Peter CY and Xu, Qingsong and Li, Zhengguo},
  journal={IEEE Transactions on Instrumentation and Measurement},
  pages={1--16},
  year={2023}
}

@article{wang2024anisotropic,
  title={Anisotropic Scale-Invariant Ellipse Detection},
  author={Wang, Zikai and Zhong, Baojiang and Ma, Kai-Kuang},
  journal={IEEE Transactions on Image Processing},
  year={2024}
}

@article{bellavia2011improving,
  title={Improving Harris corner selection strategy},
  author={Bellavia, Fabio and Tegolo, D and Valenti, Cf},
  journal={IET Computer Vision},
  pages={87--96},
  year={2011}
}

@article{bellavia2022harrisz+,
  title={HarrisZ\textsuperscript{+}:Harris corner selection for next-gen image matching pipelines},
  author={Bellavia, Fabio and Mishkin, Dmytro},
  journal={Pattern Recognition Letters},
  pages={141--147},
  year={2022}
}

\begin{IEEEbiography}[{\includegraphics[width=1in,height=1.25in,clip,keepaspectratio]{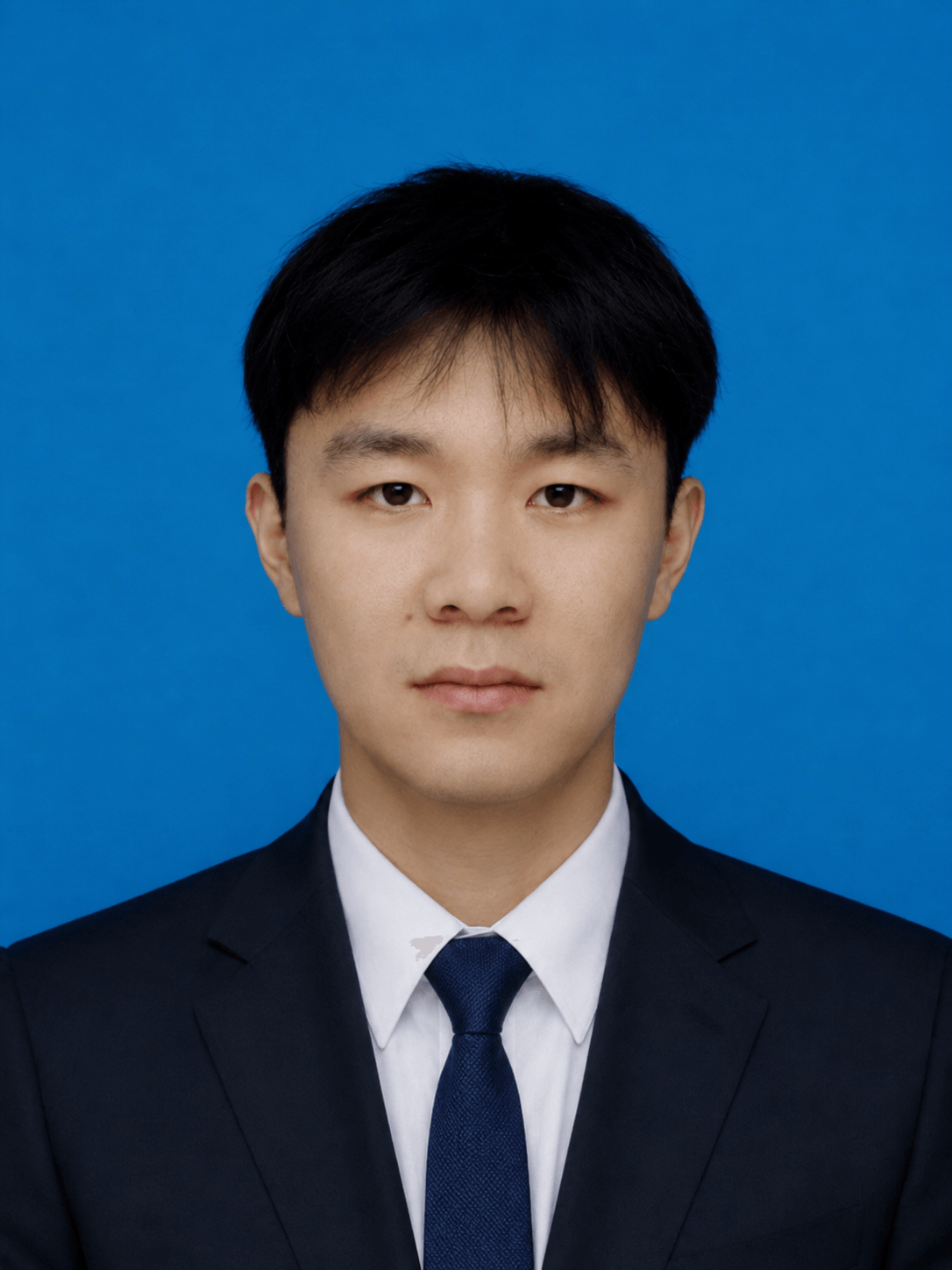}}]{Jiamiao Lu}
	received his Bachelor of Engineering degree in Electronic Information Engineering from Shaanxi University of Science and Technology in 2024 and is currently pursuing an M.Sc. in New Generation Electronic Information Technology at the School of Electronic Information and Artificial Intelligence, same university. His research focuses on computer vision and image processing, emphasizing interest point detection and few-shot learning.
\end{IEEEbiography}
\vspace{-10 mm}

\begin{IEEEbiography}[{\includegraphics[width=1in,height=1.25in,clip,keepaspectratio]{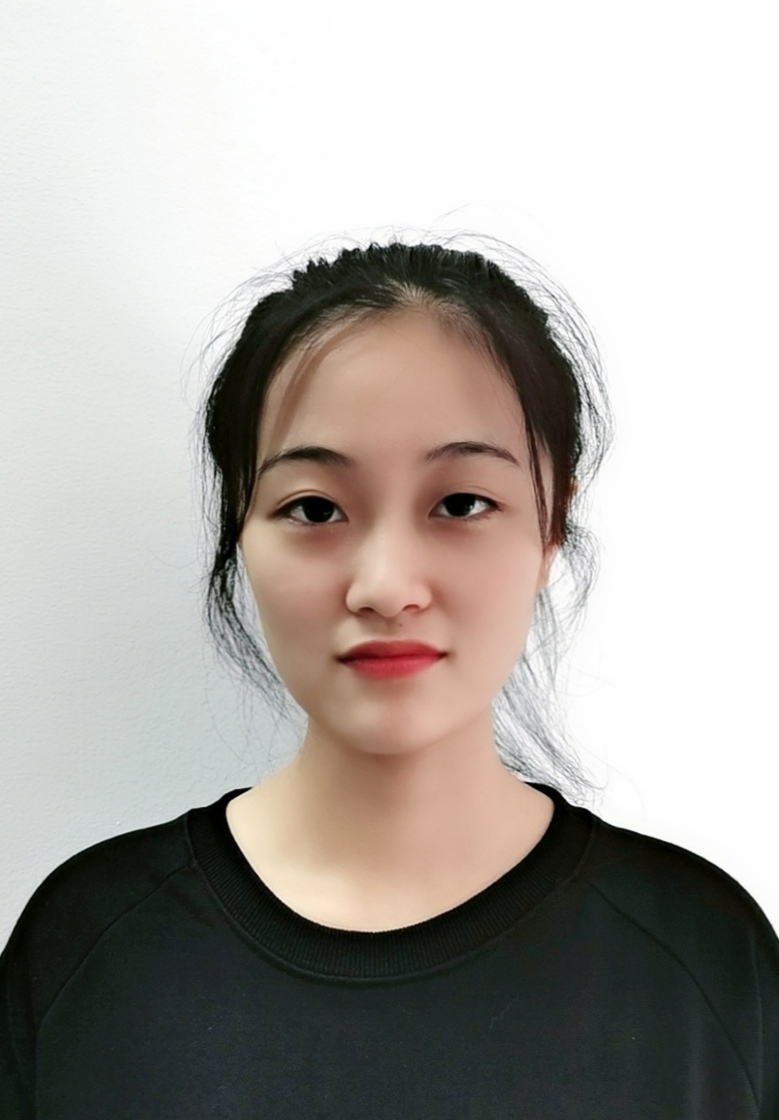}}]{Junjie~Qiu}
	received the Bachelor degree in Science from Huaiyin Institute of Technology. She is currently a M.Sc. student in Computer Technology at the School of Electronic Information and Artificial Intelligence, Shaanxi University of Science and Technology. Her research interests include computer vision and image processing, particularly in the field of interest point detection.
\end{IEEEbiography}
\vspace{-10 mm}

\begin{IEEEbiography}[{\includegraphics[width=1in,height=1.25in,clip,keepaspectratio]{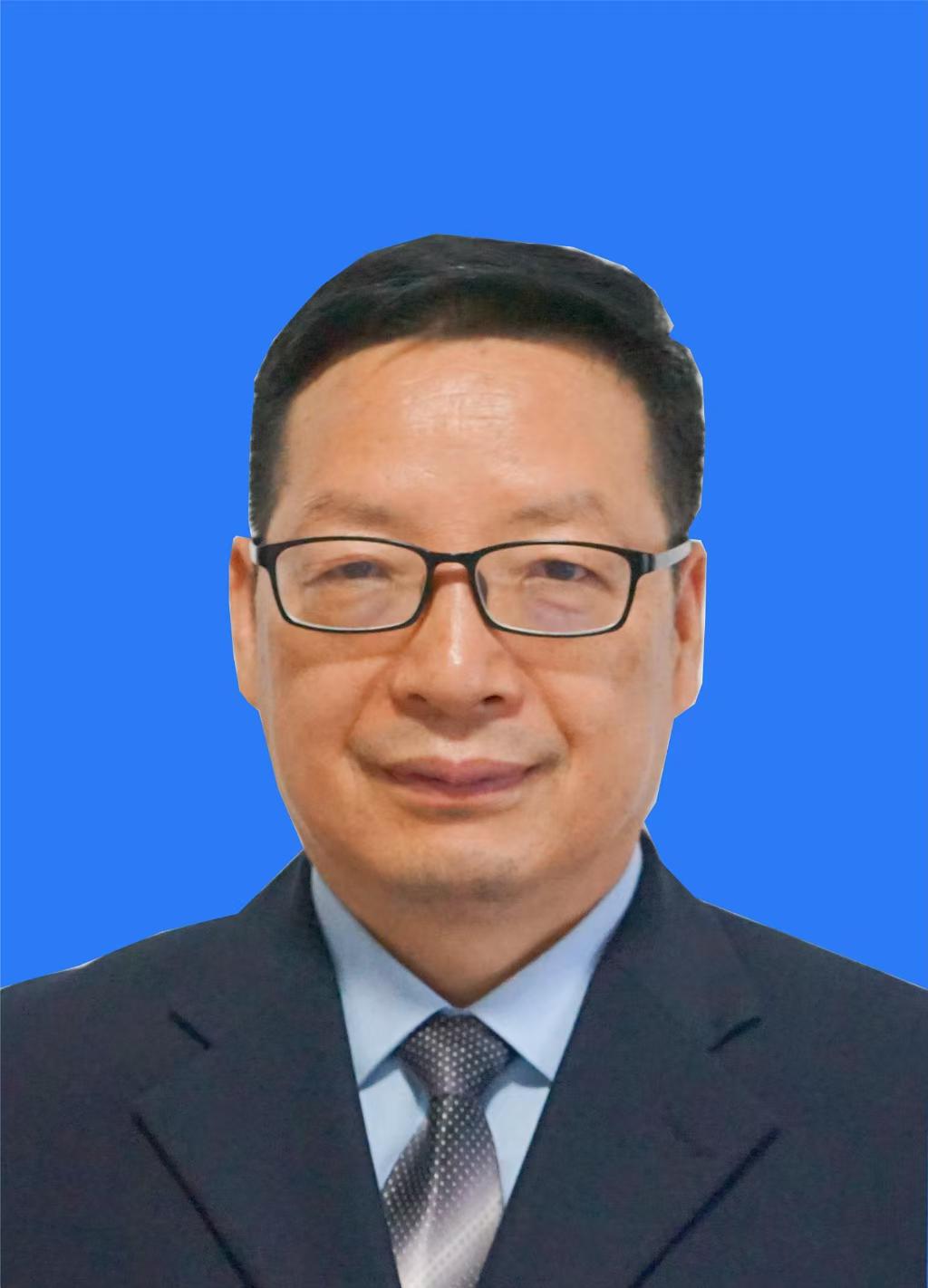}}]{Linkun Ma}
	received the B.Sc. degree in radio technology, the M.Sc. degree in transportation planning and management, and the Ph.D. degree in information and communication engineering from Northwestern Polytechnical University, China. He is currently a professor at the school of Electronic Information and Artificial Intelligence, Shaanxi University of Science and Technology, Xi'an, Shaanxi Province, and serves as a standing committee member of the ACM Xi'an Chapter. His main research interests are adaptive signal processing, implementation techniques, and Artificial Intelligence.	
	
\end{IEEEbiography}
\vspace{-10 mm}

\begin{IEEEbiography}[{\includegraphics[width=1in,height=1.25in,clip,keepaspectratio]{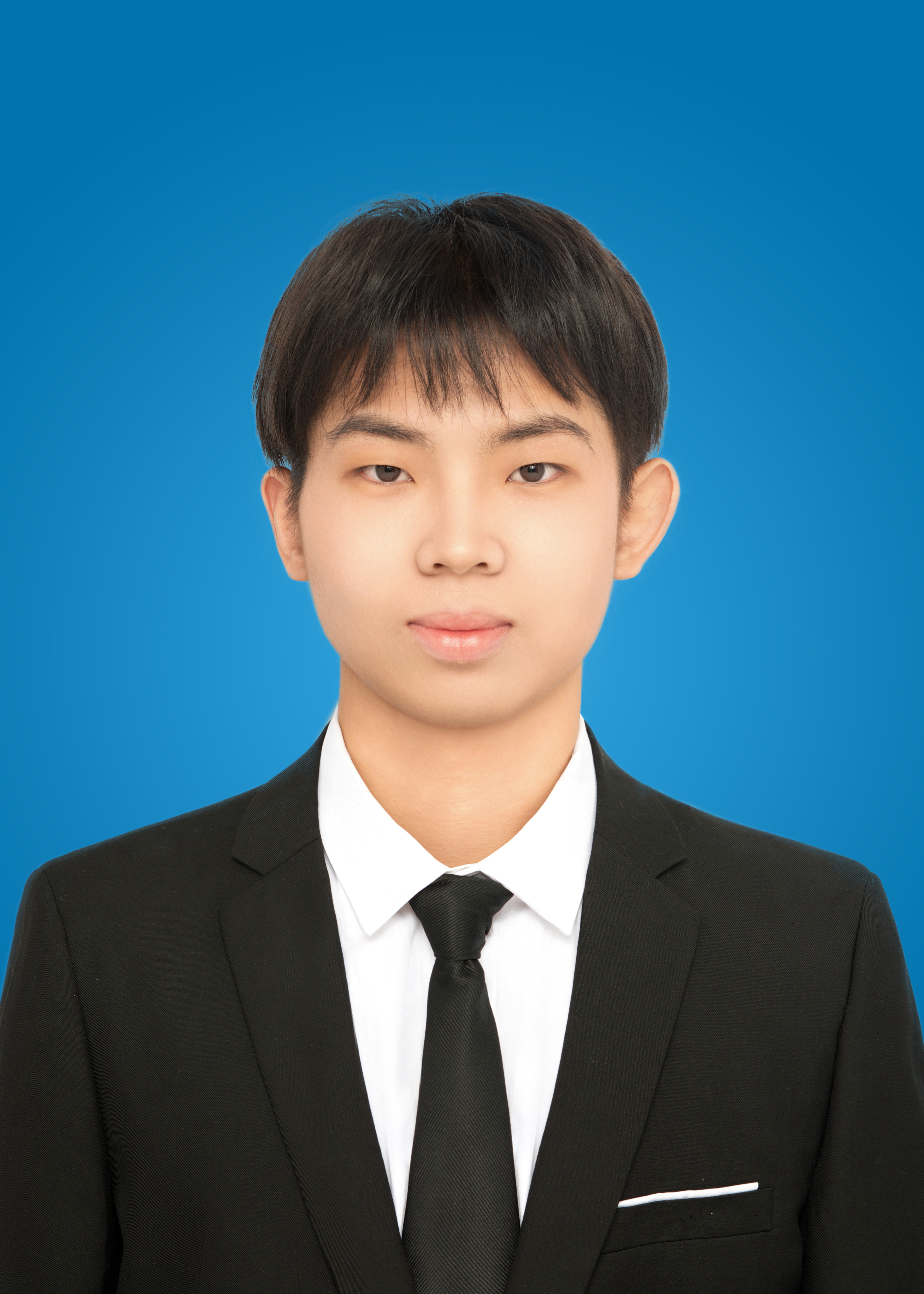}}]{Dongbo~Xie}
	received the Bachelor of Engineering degree in Computer Science and Technology from Taiyuan Institute of Technology in 2024. He is currently an M.Sc. student in Computer Technology at the School of Electronic Information and Artificial Intelligence, Shaanxi University of Science and Technology. His research interests include computer vision and image processing, particularly in the field of interest point detection.
\end{IEEEbiography}
\vspace{-10 mm}

\begin{IEEEbiography}[{\includegraphics[width=1in,height=1.25in,clip,keepaspectratio]{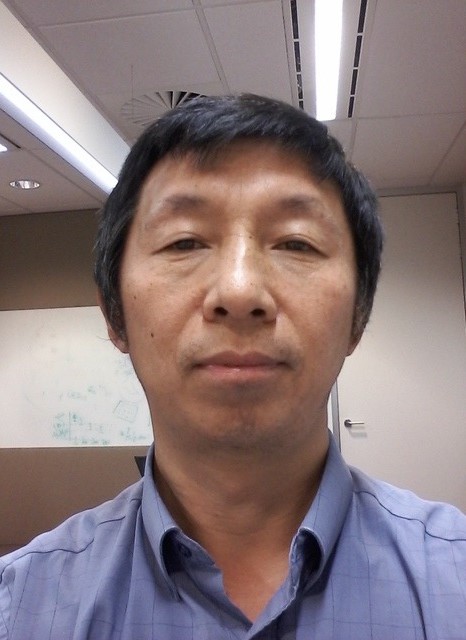}}]{Changming Sun}
	received his PhD degree in computer vision from Imperial College London, London, UK in 1992. He then joined CSIRO, Sydney, Australia, where he is currently a Principal Research Scientist carrying out research and working on applied projects. He is also a Conjoint Professor at the School of Computer Science and Engineering of the University of New South Wales. He has served on the program/organizing committees of various international conferences. He is an Associate Editor of the EURASIP Journal on Image and Video Processing. His current research interests include computer vision, image analysis, and pattern recognition.
\end{IEEEbiography}
\vspace{-10 mm}
\begin{IEEEbiography}[{\includegraphics[width=1in,height=1.25in,clip,keepaspectratio]{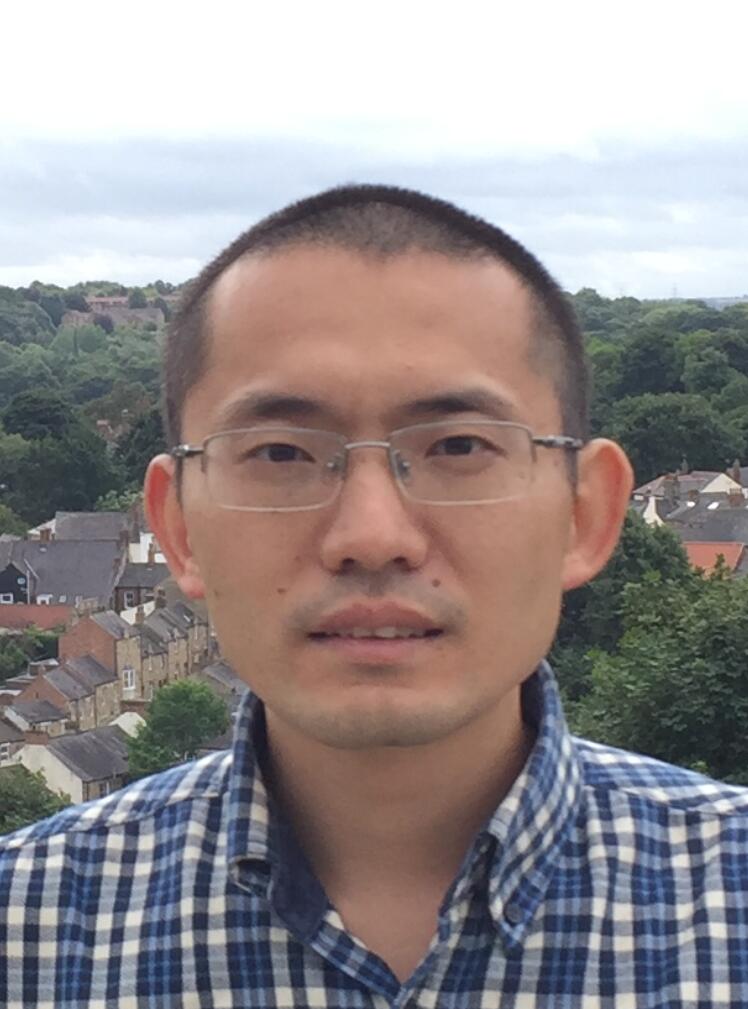}}]{Weichuan Zhang}
	received the MS degree in signal and information processing from the Southwest Jiaotong University in China and the PhD degree in signal and information processing in National Lab of Radar Signal Processing, Xidian University, China. He is currently a Professor with the School of Electronic Information and Artificial Intelligence, Shaanxi University of Science and Technology, Xi 'an, Shaanxi Province, China, and a Principal Research Fellow at Griffith University, QLD, Australia. His research interests include computer vision, image analysis, and pattern recognition. He is a member of the IEEE.
\end{IEEEbiography}

\vfill

\end{document}